\documentclass[journal,twocolumn]{IEEEtran}
\usepackage{amsmath}
\usepackage[utf8]{inputenc}
\usepackage{longtable}
\usepackage[section]{placeins}
\usepackage{multirow}
\usepackage{wrapfig}
\usepackage{makecell}
\usepackage{lscape}
\usepackage{caption}
\usepackage[numbers]{natbib}
\ifCLASSINFOpdf
\usepackage[pdftex]{graphicx}
\else
\fi
\usepackage{array}
\usepackage{stfloats}
\usepackage{url}
\begin{document}
%
\title{A Comprehensive Review of Sign Language Recognition: Different Types, Modalities, and Datasets}
%
%
%
\author{M. MADHIARASAN$^{1*}$, Member,~IEEE~\IEEEmembership{$^{1*}$Department of Computer Science and Engineering, Indian Institute of Technology Roorkee,}
PARTHA~PRATIM~ROY$^{2}$, Member,~IEEE ~\IEEEmembership{$^{2}$Department of Computer Science and Engineering, Indian Institute of Technology Roorkee}
\thanks{Dr. M. MADHIARASAN was with the Department of Computer Science and Engineering, Indian Institute of Technology Roorkee, Roorkee, Uttarakhand, India – 247667, $^{*}$Corresponding Author, e-mail: $^{1*}$mmadhiarasan89@gmail.com, mmadhiarasan.cse@sric.iitr.ac.in, $^{1*}$ Orcid ID: 0000-0003-2552-0400.}
\thanks{Dr. PARTHA~PRATIM~ROY was with the Department of Computer Science and Engineering, Indian Institute of Technology Roorkee, Roorkee, Uttarakhand, India – 247667,  e-mail:partha@cs.iitr.ac.in, Orcid ID: 0000-0002-5735-5254.}
\thanks{Manuscript received April xx, 2022; revised xx, 2022.}}
%
%
\markboth{Journal of \LaTeX\ Class Files,~Vol.~xx, No.~x, April~2022}%
{Shell \MakeLowercase{\textit{Madhiarasan et al.}}: A Comprehensive Review of Sign Language Recognition: Different Types, Modalities, and Datasets for IEEE Communications Society Journals}
%
\maketitle
\begin{abstract}
A machine can understand human activities, and the meaning of signs can help overcome the communication barriers between the inaudible and ordinary people. Sign Language Recognition (SLR) is a fascinating research area and a crucial task concerning computer vision and pattern recognition. Recently, SLR usage has increased in many applications, but the environment, background image resolution, modalities, and datasets affect the performance a lot. Many researchers have been striving to carry out generic real-time SLR models. This review paper facilitates a comprehensive overview of SLR and discusses the needs, challenges, and problems associated with SLR. We study related works about manual and non-manual, various modalities, and datasets. Research progress and existing state-of-the-art SLR models over the past decade have been reviewed. Finally, we find the research gap and limitations in this domain and suggest future directions. This review paper will be helpful for readers and researchers to get complete guidance about SLR and the progressive design of the state-of-the-art SLR model.
\end{abstract}
%
\IEEEpeerreviewmaketitle
\section{Introduction}
\IEEEPARstart
According to the WHO (World Health Organization) report, over 466 million people are speech or hearing impaired, and 80\% of them are semi-illiterate or illiterate \cite{DeafnessHearingLoss}. Non-verbal manner conveys and communicates our views, emotions, and thoughts visually through sign language. Compared to spoken language, sign language grammar is quite different. A sign comprises specific hands, shapes, or signals produced in a particular location on or around the signer’s body combined with a specific movement. \\ 
\par
\begin {figure*}
\includegraphics[width=1.0\linewidth,height=0.4\textheight]{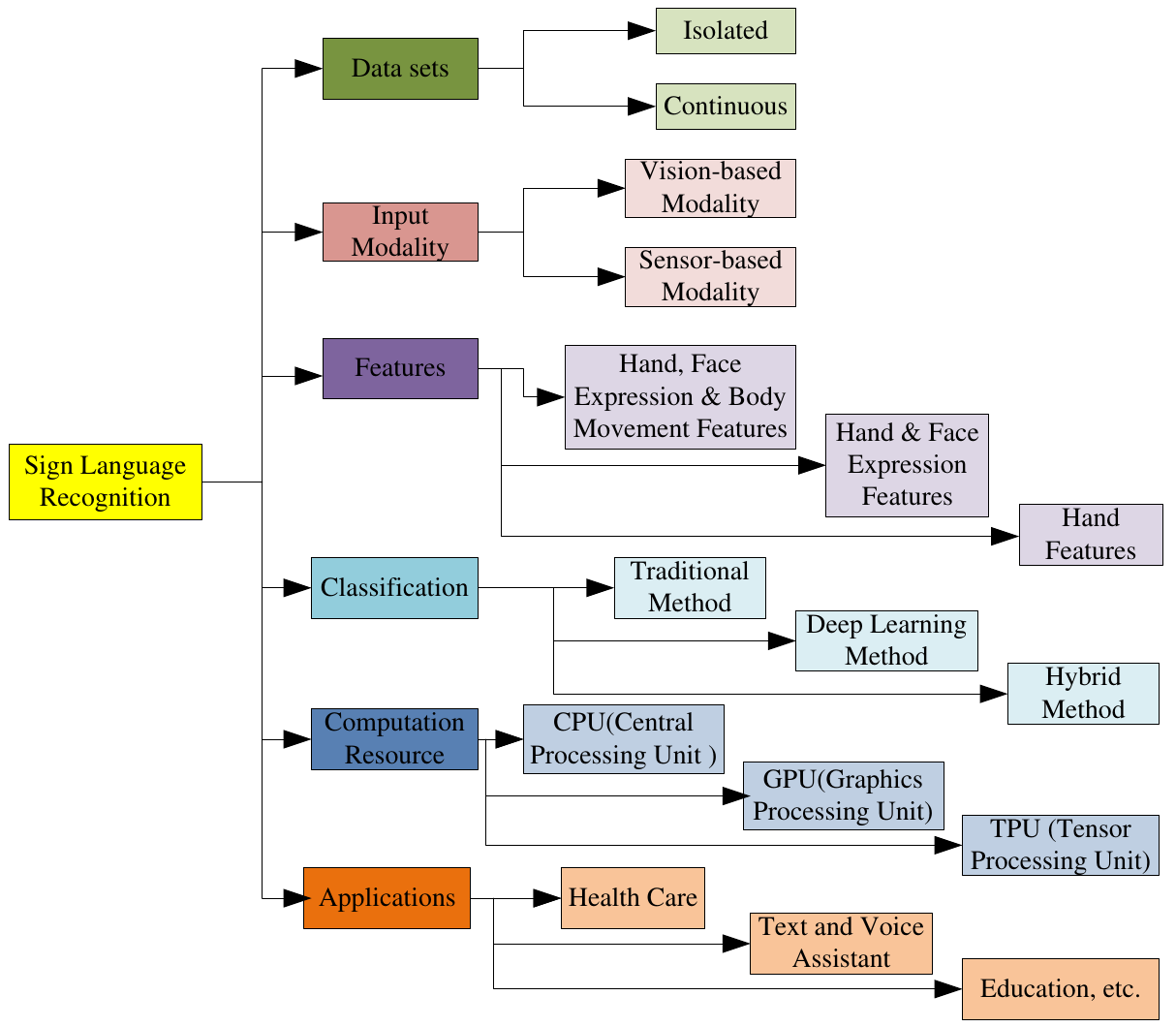}
\caption{SLR Taxonomy: the fundamental attributes of SLR like datasets, modality, features, classification, computation resources, and application, along with each attribute's categorization are shown here. Large-scale datasets and modalities affect the recognition performance. The efficient features extraction method and classification model with efficient power computation resources lead to high performance.}
\label{Figure 1}
\end{figure*}

 Hand gestures, signals, body movements, facial expressions, and lip movements are the visual means of communication used by the hand-talk community and ordinary people to convey the meaning; We recognize this language as a sign language. Sign language recognition (SLR) is challenging and complex, and many research opportunities are available with the present technology of artificial intelligence. A taxonomy of SLR is shown in Figure \ref{Figure 1}. It comprises datasets, input modality, features, classification, computational resources, and applications. The dataset is further classified into isolated sign dataset and continuous sign dataset. Vision-based modality and sensor-based modality are the general types of input modality. Hand movement, facial expression, and body movement are the major features that concern SLR. Classification is typified into traditional methods (HMM, RNN, etc.), deep learning (CNN), and hybrid method (combination of traditional and deep learning or combination of deep learning and optimization algorithm).\\
\par SLR aims to understand the gestures by suitable techniques, which requires identifying the features and classifying the sign as gesture recognition. In the literature, there is no comprehensive review paper addressing the aspect of the modality (vision and sensor), different types (isolated (manual and no manual) and continuous (manual and no manual)), various sign language datasets, and state-of-the-art methods based studies. This review study focuses on SLR-based research work, recent trends, and barrier concerns to sign language. Different sign languages, modalities, and datasets in sign language have been discussed and presented in tabular form to understand better. From databases like IEEE explore digital library, science direct, springer, web of science, and google scholar, we used the keywords sign language recognition to identify significant related works that exist in the past two decades have included for this review work. We excluded papers other than out-of-scope sign language recognition and not written in English. 
The contributions to this comprehensive SLR review paper are as follows:
\begin{itemize}
\item Carried out a review of the past two decades of published related work on isolated manual SLR, isolated non-manual SLR, continuous manual SLR, and continuous non-manual SLR.
\item Discussed different sensing approaches for sign language recognition and modality 
\item This paper presents SLR datasets concerned with isolated and continuous, various sign languages, and the complexity of the datasets discussed.
\item Discussed the framework of SLR and provided insightful guidance on SLR 
\item Point out the limitations related to the dataset and current trends available in the SLR and potential application of SLR with human-computer interaction.
\item This paper studied the results of the current state-of-the-art SLR model regarding the various benchmark SLR datasets for isolated and continuous SLR.
\item This paper analyzes current SLR issues and advises future SLR research direction.
\end{itemize}
\subsection{Need of SLR} 
As per WHO statistics, around 5\% of the population in the world suffers from a lack of hearing power. According to the prediction of the United Nations, the number of deaf people in 2050 will be 900 million \cite{DeafnessHearingLoss}. Hence, SLR receives a lot of attention at present. SLR can eliminate the communication gap between the hand-talk community (deaf and dumb). Also, SLR helps to improve communication in the following ways.
\begin{itemize}
\item It reduces the frustration of the hand-talk community.
\item The communication barrier overcome by SLR leads to effective communication.
\end{itemize}
Much research endeavored to develop high-performance SLR. Despite that, it is challenging, and it is one of the recent research fields with enormous research scope. 
\subsection{Challenges} 
SLR comprises numerous gestures and facial expressions, making it complex and challenging. In addition, to the manual components, lip shapes and eyebrow positions distinguish similar signs; e.g., many manual signs seem to be of a similar pose. However, these can be differentiated with the help of facial expression and lip movement. Sign language comprises hand movement, shape, position, orientation, palm posture, finger movement, facial expression, and body movements. These components highly influence the performance of SLR. Some of the barriers and problems of SLR are tabulated in Table \ref{Table 1}. With the advance of hardware, efficient algorithms can improve the processing speed. The scaling and image orientation problems can be resolved with recent deep learning techniques. The illumination problem can be overcome if the RGB is converted to HSV (Hue Saturation Value) or Ycbcr (Luminance Chrominance). Dynamic and non-uniform background environment problems could be resolved using the skin region and background subtraction method.
\begin{table*}[ht]
\centering
\caption{SLR Barrier, and Problem: We discussed how the barriers (dynamics, illumination of lights and environment, scaling, and computation time) of SLR cause a problem.}
\label{Table 1}
\begin{tabular}{|p{6cm}|p{11.3cm}|}
\hline
\textbf{Barrier} & \textbf{Problem} \\
\hline
Computation speed and time & Create complexity to the system and take a lot of computation time.\\
\hline
Scaling and image orientation problem &	The distance of input data capturing various signers. \\
\hline
Illumination of light &	Performance varies with different illumination scenarios because most models use the RGB model. It is highly illumination sensitive. \\
\hline
Dynamic and non-uniform background environment & The noise, improper detection of hand, and face lead to affect the performance and mislead the sign recognition system. \\
\hline
\end{tabular}
\end{table*}
\begin {figure*}
\includegraphics[width=\textwidth]{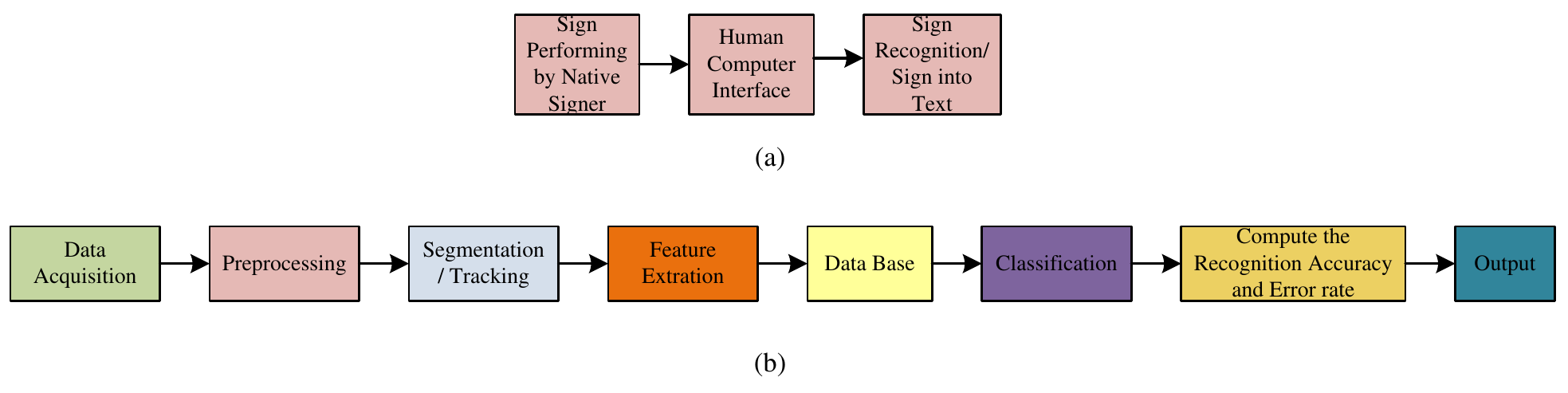}
\caption{(a) SLR (b) General process flow of SLR. Figure (a) illustrates the native signer performing sign conversion into text with the help of a human-computer interface, and figure (b) illustrates the procedural stages of SLR. The recognition rate highly depends on the data set, preprocessing, feature extraction, and classifier.}
\centering
\label{Figure 2}
\end{figure*}
\subsection{Procedure involved in SLR}
\par The SLR involves data collection, preprocessing, feature extraction, and classification phase. The block diagram of SLR and its general process is demonstrated in Figure \ref{Figure 2}. These stages are discussed in the following. Note that, for the sensor-based approach, preprocessing and segmentation are optional.
\\ \par \textbf {Data Collection:} In SLR, the data acquisition is performed using one of two modes; Vision and Sensor. In a vision-based approach, the input is an image or video \cite{aly2020deeparslr}, \cite{rastgoo2020hand}. A single camera is used to collect standard signs while multiple cameras, active and invasive devices, help collect the depth information. Video camera, webcam, or smartphone device \cite{tripathi2015continuous}, \cite{hassan2019multiple}, \cite{albanie2020bsl}, \cite{aran2009signtutor} captured the continuous motion. The sensor-based approach collects the signal with the use of the sensor \cite{botros14electromyography}, \cite{fatmi2019comparing}, \cite{wei2016component}, \cite{kim2018finger}. \\ 
\par \textbf{Image Preprocessing:} The performance of the SLR system can be improved by preprocessing methods such as dimension reduction, normalization, and noise removal. \cite{cheok2019review}. \\
\par \textbf{Segmentation:} The segmentation stage splits the image into various parts or ROI (Region of Interest) \cite{kim2018effective}, Skin Colour Segmentation \cite{paulraj2010phoneme}, HTS (Hands Tracking and Segmentation) \cite{ghotkar2013vision}, Entropy Analysis and PIM (Picture Information Measure) \cite{shin2006hand}. The background requires the hand gesture extraction to be done effectively by segmentation and tracking process.\\ 
\par \textbf{Tracking:} Tracking of hand position and facial expression from the acquired image/video can be performed using camshaft (continuously adaptive mean shift used to track the head position) \cite{akmeliawati2009towards}, Adaboost with HOG (Histogram of the gradient) \cite{wang2012hidden}, Particle filtering (KPF-Kalman Particle Filter) \cite{li2003visual}.\\ 
\par \textbf{Feature Extraction:} Transforming preprocessed input data into the feature space is known as feature extraction. Further, it is discussed in detail in section 2.\\ 
\par \textbf{Data Base:} The acquired data (image/video) is stored in the database and classified into two sets, namely training and testing datasets \cite{kadhim2020real}. The classifier learns by training dataset and the performance is evaluated by testing data.\\ 
\par \textbf{Classification:} The classifiers perform the classification by extracting features and classify the sign gesture. The Hidden Markov Model (HMM) \cite{fatmi2019comparing}, \cite{forster2013modality}, Long-Short Term Memory (LSTM) \cite{lee2021american} Deep Learning network \cite{al2020deep}, and hybrid classifier \cite{aly2020deeparslr}, \cite{yuan2020hand} are used as classifiers to recognize sign language.\\ 
\par \textbf{Evaluation Stage:} The performance of a trained classifier is validated with a testing dataset (unseen data during training) \cite{vamplew1996recognition}. The error incurred during classification gauges sign recognition performance. \\

\par Although there are few review papers in the literature \cite{kudrinko2020wearable}, \cite{cheok2019review}, however, they lack focus and understanding of SLR. This paper provides a comprehensive SLR preamble, recent research progress, barriers or limitations, research gap, and future research direction and scope.
We organized the rest of the review paper as follows. Section 2 presents sign language modality, preprocessing, and the various feature extraction methods in SLR. Carried out a literature review concerning the manual and non-manual aspects of SLR in Section 3; Section 4 discusses and illustrates the classification architecture of SLR. Section 5 presents various types of SLR, datasets concerning SLR, and reviews work related to the modalities, current state-of-the-art models based on SLR. The recent trends, challenges, and limitations are highlighted in Section 6. Sections 7 and 8 pointed out future research discussion and conclusion, respectively.
\section{Modalities of SLR} 
\begin {figure}
\centering
\includegraphics [width=0.6\linewidth,height=0.25\textheight] {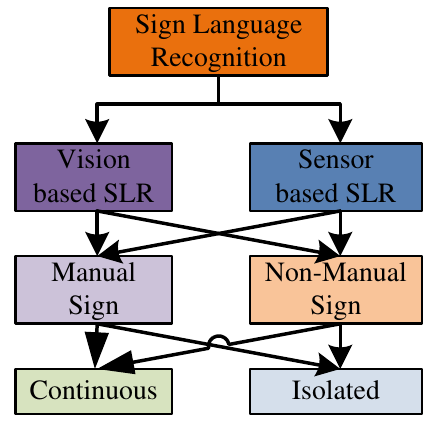}
\caption{SLR Types: Vision-based SLR and sensor-based SLR are the SLR types. It is further, classified into manual and non-manual, then isolated and continuous.}
\centering
\label{Figure 3}
 \end{figure}

SLR is one of the most prominent research areas in computer vision and natural language processing. In concern to the acquisition process, the SLR system is classified as a sensor-based and vision-based approach. Both approaches are next classified as manual and non-manual, and further classified as isolated and continuous. Figure \ref{Figure 3} illustrates the SLR types. Much research work focused on isolated manual-based SLR. Only a little research work addressed continuous non-manual SLR.\\
\begin{figure}
\centering
\includegraphics[scale=.62]{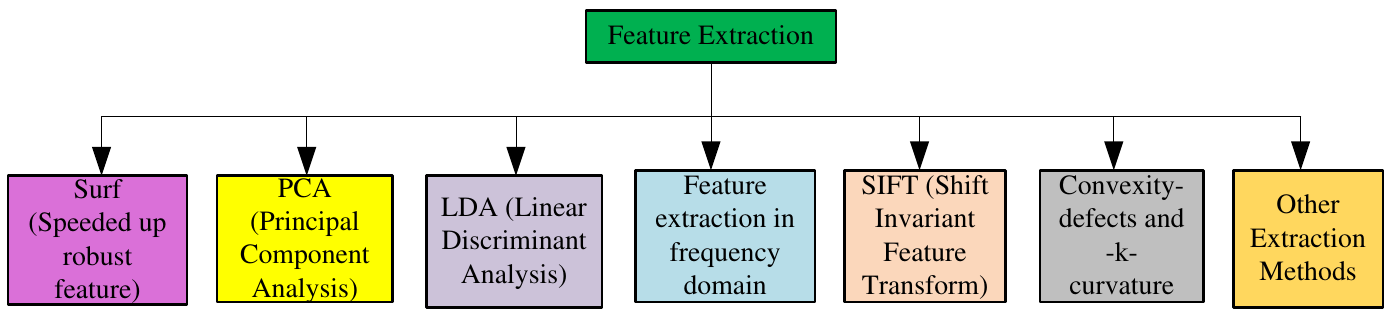}
\caption{Various methods are used to extract the significant features.}
\centering
\label{Figure 4}
\end{figure}
\begin{table*}[ht]
\centering
\caption{The importance of vision-based and sensor-based methods are shown here according to the obstacle, cost, merits, and demerits. Much research work focuses on vision-based SLR because of its feasibility in real-time applications.}
\label{Table 2}
\begin{tabular}{|p{1.5cm}|p{2cm}|p{2cm}|p{2.5cm}|p{0.5cm}|p{4cm}|p{2.5cm}|}
\hline
\textbf{Method} &	\textbf{Capturing Device} & \textbf{Obstacle} & \textbf{Efficiency}	&	\textbf{Cost} & \textbf{Limitation} & \textbf {Advantage}\\
\hline
Vision-based Method & Video Camera & Environment, disturbance, and noise. & Low (depends on the resolution). & Low & Possess challenging concerns for time, speed, and overlapping. More Feature extraction techniques are required. & Fast Speed.\\
\hline
Sensor-based or gloves-based Method & Sensors and gloves &	Environment, disturbance, and noise. & Better than vision-based method (depends on sensor performance). & High & Not suitable for real-time application. & Better performance. Require minimal feature extraction.\\
\hline
\end{tabular}
\end{table*}
\begin{table}
\centering
\caption{Various features extraction methods associated with SLR work exist in the literature.}
\label{Table 3}
\begin{tabular}{|p{4.2cm}|p{0.5cm}|p{2.7cm}|}
\hline
\textbf{Method} & \textbf{Year} & \textbf{Author}\\
\hline
FPM (Feature Pooling Module) &	2020 & \citet{sincan2020autsl}\\
\hline
Histograms of oriented gradients &	2016 &\citet{chansri2016hand}\\
\hline
\multirow{2}{*} {Euclidean distance}	& 2016 & \citet{pansare2016vision}\\
\cline{2-3}
&  2013 &	\citet{singha2013indian}\\
\hline
\multirow{2}{*} {DWT (Discrete Wavelet Transform)} &	2017 & \citet{ahmed2016vision}\\
\cline{2-3}
& 2016 & \citet{prasad2016indian}\\
\hline
SIFT (Scale Invariant Feature Transform) &	2012 &	\citet{Gurjal2012indian} \\
\hline
SURF (Speeded Up Robust Feature) &	2012 &	\citet{yao2012hand}\\
\hline
\multirow{2}{*}{Fourier Descriptors}	& 2017 &
\citet{kumar2017sign}\\
\cline{2-3}
& 2015	 & \citet{shukla2015dtw}\\
\hline
\multirow{2}{*}{PCA (Principal Component Analysis) }&	2021 &	\citet{gurbuz2021american}\\ \cline{2-3}
&	2015 &	\citet{tripathi2015continuous}\\
\hline
Fuzzy neural network &	2016 &	\citet{dour2016recognition}\\
\hline
\end{tabular}
\end{table}
\par \textbf{Sensor-based approach:} Physically attached sensors acquire trajectories of the head, finger, and motion of the signer. Sensor-associated gloves track the signer’s hand articulations and recognize the sign. The comparison of SLR methods shown in Table \ref{Table 2} clarifies vision and sensor-based approaches. In contrast with vision-based SLR, sensor-based SLR provides efficient performance.\\
\par \textbf{Vision-based approach:} The gestures captured by multiple cameras (or webcam) are recognized using the vision/image-based approach. From the captured image/video, it extracts palm, finger, and hand movement features. With the help of these extracted features, classification was performed. Poor illumination or lighting environment, noisy background, and blurring present in the image result in misclassification. Although vision-based SLR is suitable for real-time conditions, it must adequately care for preprocessing, feature extraction, and classification.
\subsection{Preprocessing}
The computational burden of data processing could be reduced by preprocessing methods. Image reduction and image conversion methods do the size reduction and conversion from color to gray scale. Image reduction methods reduce the burden of data processing. The unwanted object can be removed by the histogram equalization \cite{sethi2012signpro}. The noise present in the image are removed using the filter, like median, moving average method, and so on \cite{lahiani2015real}. Gaussian average methods are used to remove the image background component \cite{pansare2012real}. Filters perform removal of the unwanted components and minimize the size of the data with the help of image edge detection algorithm \cite{lionnie2012performance}. The filter process speeds up with the help of fast Fourier transformation because instead of an image, the frequency domain is used \cite{zhang2016recognition}. The image is split into possible segments \cite{zorins2016review}; masking is used in segmentation to improve processing. Elimination of background effect using binarization histogram equalization aid for better image contrast. Normalization methods can effectively handle the variance in the data \cite{tsagaris2013colour}.
\subsection{Feature Extraction}
\par In SLR, relevant feature extraction plays a vital role. It is crucial for sign language, as irrelevant features lead to misclassification \cite{al2020hand}. The feature extraction aid in accuracy improvement, and speed \cite{khalid2014survey}. Some of these feature extraction method include SURF (Speeded Up Robust Feature) \cite{yao2012hand}, speed up robust feature (Laplace of Gaussian with box filter) \cite{yao2012hand}, SIFT (shift-invariant feature transform) \cite{Gurjal2012indian}, PCA (Principal Component Analysis) \cite{gurbuz2021american}, \cite{tripathi2015continuous}, LDA (Linear Discriminant Analysis) \cite{jiang2018feasibility}, Convexity defects and k-curvature \cite{tariq2012sign}, time domain to frequency domain \cite{ahmed2016vision}, \cite{kumar2017sign}, Local binary pattern, etc. The feature extraction methods used for SLR-based study is tabulated in Table \ref{Table 3}. Various feature extraction methods are showed in Figure \ref{Figure 4}. Feature vector dimension reduction performed by PCA, LDA, etc. aid in reducing the computational burden on the classifiers. The dimensionality pruning, features reduction, and lowering of the dimension keep the significant features of high variance and minimizing remaining features, thus, reduces the training complexity. Fourier descriptors are noise resistance and invariant to scale, orientation, and normalization is easy. The process of transforming the correlated into an uncorrected value is known as principal component analysis. Original data are linearly transformed effectively, and the feature vectors get reduced. \\

\par The preprocessing and feature extraction methods aid the classifier. Also, they reduce the computation burden, avoid overfitting issues, and wrong recognition possibilities. SIFT's merits are invariant to lighting, orientation, and scale. However, the performance is not satisfactory \cite{Gurjal2012indian}. Using Histogram of Oriented Gradients (HOG) \cite{chansri2016hand}, the unwanted information is removed, keeping the significant features to ease the image processing. The feature vectors are obtained using the computation of gradient margin and angle. As HOG cell size and the number of bins increase, the extracted feature also increases. Larger subdivisions furnish global information, and small subdivisions given local information that is worthwhile. The demerits of both the methods are that they require more memory. SURF is invariant to image transformation and a faster feature extractor than SIFT. Still, it has the requirement of camera setup in horizontal position for better performance, and the disadvantage is illumination-dependent, not rational. Location and frequency captured using a Discrete Wavelet Transform. Temporal resolution is the critical merit of DWT \cite{ahmed2016vision}, \cite{prasad2016indian}.

\section{Literature studies about SLR} Sign language is not generic; it varies according to the region and country \cite{DeafnessHearingLoss}. The sign language classification is available in over 300 sign languages worldwide, namely ASL, BSL, ISL, etc. According to Ethnologue 2014 \cite{ethnologue} in the United States, ASL is a native language for around 2,50,000-5,00,000 people. Chinese Sign Language is being used in China by approximately 1M to 20M deaf people. Approximately 1,50,000 people in the United Kingdom use British Sign Language (BSL). In Brazil, approximately 3 million signers use Brazilian Sign Language to communicate, like Portuguese Sign Language or French Sign Language. According to Ethnologue 2008 in India, approximately 1.5 million signers use Indo-Pakistani Sign Language.\\

\par SLR is not only meant for deaf and mute people. Ordinary people also communicate information in the noisy area of public places and the library without disturbing others. Manuel (communication by hands) and non-manual (communication by body posture or facial expression) medium are usually used in sign language. People use sometimes finger spelling  which is communicated by splitting words into letters, then spelling the letter using fingers). Manual and non-manual SLR are discussed in detail in following subsections.

\subsection{Manual SLR} 
\begin {figure}
\includegraphics[scale= 0.72]{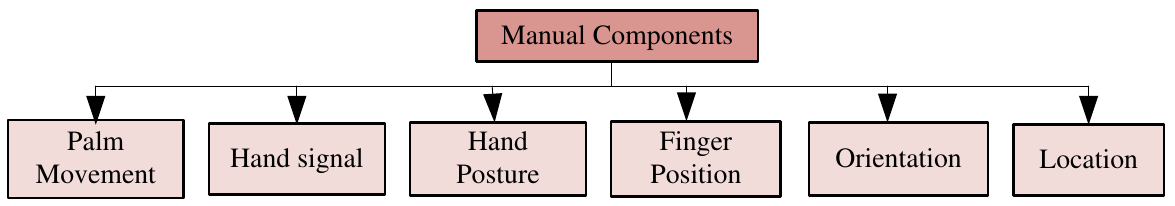}
\caption{Manual Components. The important manual features related to sign language are shown here.}
\centering
\label{Figure 5}
\end{figure}

Hand motion, hand posture, hand shape, and hand location are the manual sign components. Figure \ref{Figure 5} shows the manual sign components. With one hand or two hands, the signer usually communicates with others. The manual SLR is classified into isolated and continuous.
\subsubsection{Isolated Manual SLR}
The literature work on isolated manual SLR are as follows:
\par \textbf{Classical methods:} \citet{ong2012sign} suggested sequential Pattern Tree-based multi-class classifier for DGS (German Sign Language (Deutsche Gebärdensprache) and Greek Sign Language (GSL) recognition. Their proposed SP-Tree Boosting algorithm-based recognition model performs better than the Hidden Markov Model. \citet{chansri2016hand} proposed data fusion incurred ANN-based Thai SLR model. They extracted the hand feature using histograms of oriented gradients, and they did classification using a back-propagation algorithm associated with an ANN. \citet{yin2018research} performed hand gesture recognition using a joint algorithm based on BP and template matching method combination. The joint algorithm takes computation time as 0.0134 and an accuracy of 99.8\% was achieved for isolated hand gesture recognition. \citet{jane2018sign} carried out an ANN classifier with an association of data fusion. They performed three hidden layers of artificial neural network with wavelet denoising and TKEO (TeagerKaiser energy operator) methods for a SEE (Signing Exact English). Based on this approach, the recognition rate is 93.27\%. Korean finger language recognition model was developed based on ensemble ANN \cite{kim2018finger}. The performance was analyzed by varying dataset size (50 to 1500) and classifier (1 to 10). The comparative analysis of eight ANN classifier-associated ensemble models identifies 300 training datasets as an optimal structure to lead to 97.4 \% recognition accuracy for Korean finger language recognition. \par \citet{almeida2014feature} extracted seven vision-based features  using RGB-D sensor. They recognized Brazilian Sign Language with an average of 80\% using the SVM. They did phonological structure-based decomposition and extraction of signs. Hence, they suggested a model suitable for other SLR purposes. \citet{fatmi2019comparing} performed SLR based on ANN and SVM. They have compared their performance with HMM. Comparison with other machine learning techniques to ASL words, higher accuracy achieved by proposing ANN. \citet{lee2018smart} developed SVM based on a sign language interpretation device with 98.2 \% recognition accuracy. SVM classifier-based sign interpretation device developed system. \citet{wei2016component} presented the CSL sign recognition model using a code matching method by including a fuzzy K-mean algorithm. They determined subclass by a fuzzy K-mean algorithm and classification was done with the Code matching method. \citet{li2018skingest} suggested ASL recognition prototype model based on KNN, LDA, and SVM classifiers. They carried out a prototype model based on LDA, KNN, and SVM classifiers using a firmly stretchable strain sensor for ASL 0-9 number sign recognition. The authors reported that the model achieved an average accuracy of 98\%.  

\par     \citet{yang2017chinese} performed a Chinese Sign Language (CSL) recognition model based on sensor fusion decision tree and Multi-Stream Hidden Markov Models classifier. They developed a wearable sensor associated with the Chinese SLR model with user-dependent and user-independent using Multi-Stream Hidden Markov Models. The searching range improved by optimized tree-structure classification. 
\citet{dawod2019novel} carried out work on real-time recognition model for ASL alphabets and numbers sign recognition. They used RDF (Random Decision Forest) and HCRF (Hidden Conditional Random Field) based classifiers and Microsoft Kinect sensor v2 for the data collection. The HCRF classifier-based recognition model gets the mean accuracy for numbers-based sign recognition as 99.99\% and alphabets sign recognition as 99.9\%. The RDF-based recognition model achieved mean accuracy for number sign recognition as 96.3\% and alphabets sign recognition as 97.7\%. Hence, the HCRF based sign recognition model leads to better performance than RDF for both ASL numbers and alphabets recognition. \citet{hruz2009input} presented a Hidden Markov Model-based Czech SLR model with an association of kiosk. Also, they performed SLR, automatic speech recognition, and sign language synthesis.

\par \citet{mummadi2018real} proposed an LSF model based on IMU sensors associated with wearable hand gloves with various classifiers like naïve Bayes, MLP, and RF. Real-time wearable IMU sensor-based glove-associated sign recognition model developed for LSF recognition instead of complimentary filter advanced fusion strategy and the advanced classifier can improve the accuracy rate. \citet{botros14electromyography} presented a comparative analysis of wrist-based gesture recognition using EMG signal. Forearm and wrist level-based gesture are recognized using EMG signal. \citet{gupta2020indian} performed a wearable sensor-based multi-class label incurred SLR model. The LP-based SLR model has a minimal error and computation time than the tree-based, BR (binary relevance), and CC (Classifier Chain) based sign recognition models. Compared to the classic tree classification model, the suggested model performs well with minimal classification errors. \citet{hoang2020hgm} presented a new vision-based captured ASL alphabets sign dataset (HGM-4). With this dataset, using a classifier, developed a contactless SLR system. 

\par \textbf{Deep learning approaches:} \citet{al2020hand} performed sign dependent and sign independent SLR using three datasets using single and fusion parallel 3DCNN. The proposed model gets a better recognition rate than other considered six existing literature methods. \citet{sincan2020autsl} performed CNN and LSTM based SLR model for Turkish SLR. The feature extraction improved by FPM (Feature Pooling Module), convergence speeds up using the attention model. \citet{yuan2020hand} pointed out DCNN (deep convolution neural network and LSTM (long short-term memory) based model for hand gesture recognition. The residual module has overcome the gradient vanishing and overfitting problem. Complex hand gesture long-distance dependency problem addressed by improved deep feature fusion network. Compared to Bayes, KNN, SVM, CNN, LSTM, and CNN-LSTM, the DFFN based model performs well on ASL and CSL datasets.
\par \citet{aly2020deeparslr} designed an Arabic SLR model using BiLSTM (deep Bi-directional Long Short Term Memory recurrent neural network). Convolutional Self-Organizing Map for hand shape feature extraction, and DeepLabv3+ extracts hand regions. The suggested model proved validity on signer-independent real Arabic SLR. The proposed model is suitable for an isolated sign, and continuous sign-based analysis can be a future direction. \citet{rastgoo2020hand} carried out work on a multi-modal and multi-view hand skeleton-based SLR model. Features fusion and single-view vs. a multi-view projection of hand skeleton-based performance analysis performed. SSD (Single Shot Detector), 2DCNN (2D Convolutional Neural Network), 3DCNN (3D Convolutional Neural Network), and LSTM (long short-term memory) based deep pipe-line architectures were proposed to recognize the hand sign language automatically. \citet{lee2021american} designed the k-Nearest-Neighbour method associated with Long-Short Term Memory (LSTM) recurrent neural network-based American SLR model. The leap motion controller is used to gain the sign data. Compared to SVM, RNN, and LSTM models, the proposed model (LSTM with KNN) outperforms 99.44\%.
\par
\begin {figure}
\includegraphics[scale= 0.36]{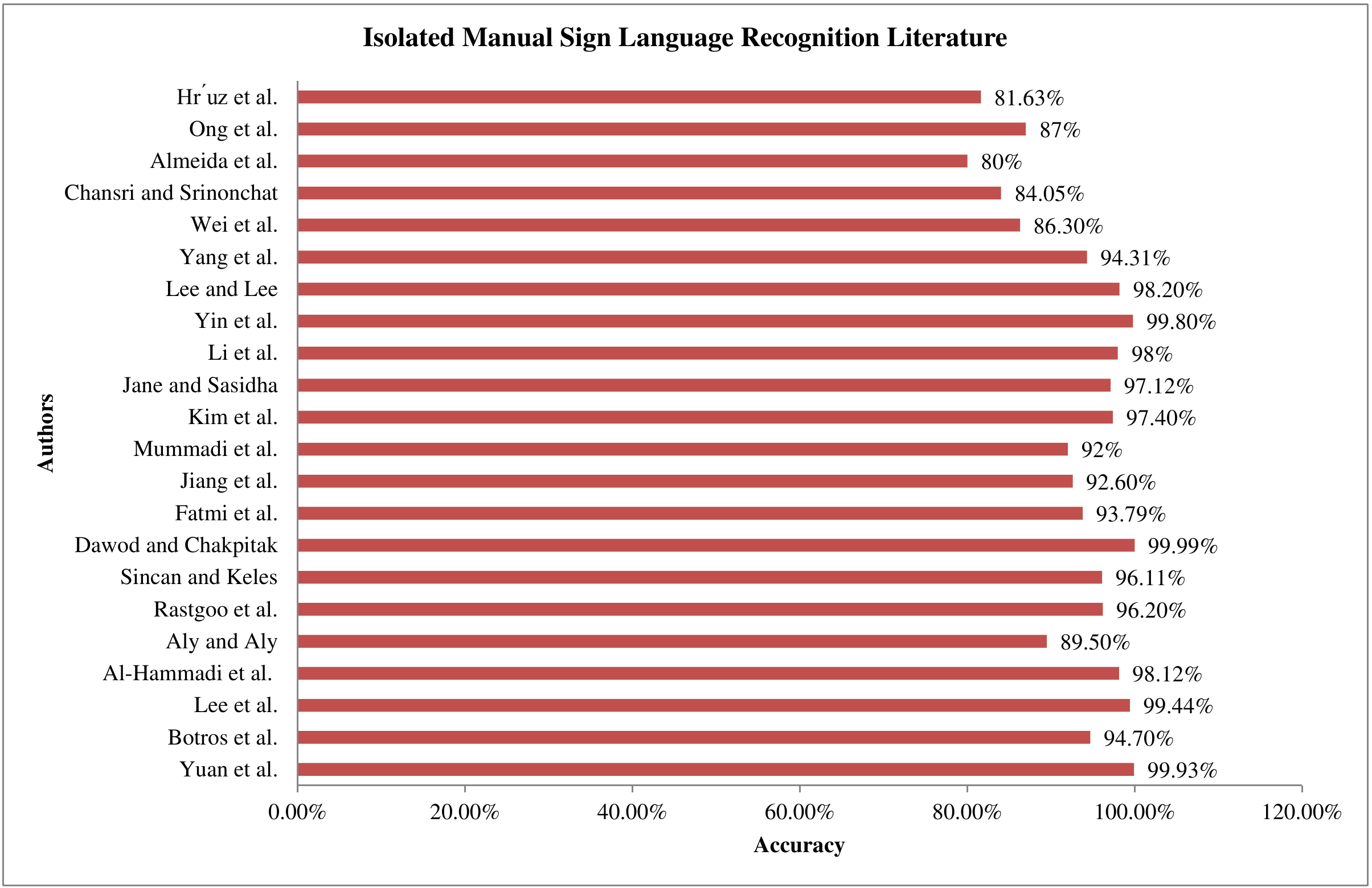}
\caption{Isolated Manual SLR Literature. This bar chart representation depicts the mean accuracy and recognition rate achieved in the literature related to the isolated manual SLR concerning different sign languages, models, and modalities.}
\centering
\label{Figure 6}
\end{figure}

 For a clear understanding, the research work related to isolated manual SLR are tabulated in Table \ref{Table 4} and graphical representation is shown in Figure \ref{Figure 6}. The recognition model results in good accuracy for isolated sign recognition, not assured to be generalized for continuous sign recognition with better precision.

\subsubsection{Continuous Manual SLR}
Processing one-dimensional data is simpler compared to handling a high-dimension dataset like video \cite{elakkiya2020machine}. Continuous SLR with uncontrolled environment-based SLR is quite complex as there is no clear pause after each gesture. 
\clearpage
\onecolumn
\begin{landscape}
\tiny
\begin{longtable}[c]{|p {0.4 cm}|p{1 cm}| p {2.5 cm}|p {2.5 cm}|p{2.5 cm}|p {3.5 cm}| p{1.5 cm}|p {1.5 cm}|p{1.7 cm}|p{2.5 cm}|}
\caption{Isolated Manual SLR Literature Work. Related work with regard to vision and sensor based SLR model concerns the isolated manual sign comprehensively summarized in a tabular form for better understanding.}
\label{Table 4}
\\\hline
\textbf{Year} &
\textbf{Author} &
\textbf{Method} &
\textbf{Pros} &
\textbf{Cons} &
\textbf{Accuracy or Result} &
\textbf{Sensor} &
\textbf{Sample and Lexicon Size} &
\textbf{Model Type} &
\textbf{Dataset}
\endfirsthead
\hline
\textbf{Year} &
\textbf{Author} &
\textbf{Method} &
\textbf{Pros} &
\textbf{Cons} &
\textbf{Accuracy or Result} &
\textbf{Sensor} &
\textbf{Sample and Lexicon Size} &
\textbf{Model Type} &
\textbf{Dataset}
\endhead
\hline
\endfoot
\endlastfoot
\hline
2021 & \citet{yuan2020hand} & CNN (Convolutional Neural Network) and LSTM (Long Short-Term Memory) & Wearable gloves based hand gesture recognition model. & Real-time live data-based analysis with the uncontrolled environment not performed to prove the validity. & Recognition accuracy: 99.93\% for ASL and 96.1\% for CSL. & 3-dimensional flex sensor-based data gloves, gyroscope, accelerometer, and bending sensor. & 26 ASL alphabets, daily activities for CSL, 6 subjects & America sign language, Chinese sign language. & A total of 5178 hand motions of ASL alphabets and 100 daily life activities sign of CSL with a total of 34452 hand motions. \\ \hline
2021 & \citet{botros14electromyography} & Linear Discriminant Analysis (LDA) used as a classifier, PCA used to reduce the dimension. & The reliable wearable device, less affected by noise & Continuous sign based experimentation was lacking limited regard to the dataset. & For multi-finger gestures, accuracy: 91.2\%, single finger accuracy: 92.1\%, and conventional wrist gestures accuracy: 94.7\%. & EMG sensor & 17 various single fingers, multi-finger, and wrist  gestures, 21 subjects &
human-computer interaction: Hand gesture sign. & 72 repetitions per participant of 5 single-finger, 6 multi-finger gestures, and 6 wrist gestures are collected. \\ \hline
2021 & \citet{lee2021american} & K-Nearest-Neighbour method, Long-Short Term Memory (LSTM), and Recurrent Neural Network. & Suitable for real-time environment/ applications. & One hand sign (right hand) only considered for recognition. & Recognition accuracy: 99.44\%  and validation (5- fold cross-validation) accuracy: 91.8\%. & Leap motion controller &
26 ASL alphabets, 100 subjects. & American SLR & 2600 samples (26 *100 samples). \\ \hline
2020 & \citet{gupta2020indian} & Multi class classification. & Computation time is less. & Expensive and continuous sign based on experimentation was lacking. & LP (label power set) model average classification error: 2.73\%. & Surface electromyogram and inertial measurement units. & 100 signs, 10 subjects. & Indian SLR & 20,000 samples. \\ \hline
2020 & \citet{hoang2020hgm} & Multi camera-based sign (ASL) data collection. & Different position of hand gesture captured & Gesture recognition not performed. & NA & 4 cameras (front, back, right, and left), laptop camera. & 26 letter sign, 5 subjects. & America sign language. & 4,160 samples. \\ \hline
2020 & \citet{al2020hand} & Single 3DCNN, and PARALLEL 3DCNN. & Better recognition rate, generalize well with three datasets. & Optimal selections of hyperparameter problem, real-time practical implementation with a live sign, was lacking. & Recognition rate for dataset 1  (SSL -SAUDI SIGN LANGUAGE) single 3DCNN signer-dependent mode: 96.69\%, signer independent recognition rate: 72.32\%, parallel 3DCNN signer-dependent mode: 98.12\%, and signer independent recognition rate: 84.38\%, for dataset 2  (ArSL-ARABIC SIGN LANGUAGE) signer-dependent mode: 100\%, signer independent recognition rate: 34.9\%, for dataset 3 (ASL-AMERICAN SIGN LANGUAGE) signer-dependent mode: 76.67\% and signer independent recognition rate: 70\%. & RGB cameras, Microsoft Kinect, analog camcorder. & Dataset 1 (40 subjects, forty gesture classes(200 gesture)), dataset 2 (3 subjects, 23 gestures, 150 samples), dataset 3 (43    classes, 40 subjects). & SAUDI SIGN LANGUAGE, ARABIC SIGN LANGUAGE, AMERICAN    SIGN LANGUAGE. & Dataset 1: signer dependent total 8000 samples, signer independent total 6400 video samples. Dataset 2: signer dependent total 3444 samples, signer independent total 3444 samples.  Dataset 2: signer dependent total 280 samples, signer independent total 280 samples. \\ \hline
2020 & \citet{aly2020deeparslr} & Convolutional SOM, deep Bi-directional LSTM network, and DeepLabv3+. & Signers independent combinational sign recognition model. & Experimentation performed with a limited number of signers. & Accuracy without segmentation: 69.0\% and with segmentation accuracy: 89.5\%. & Video camera: 25 fps (frames per second). & 23 words, 150 sequences, and 3 subjects. & Arabic sign language. & 3450 samples. \\ \hline
2020 & \citet{rastgoo2020hand} & Midpoint algorithm, SSD (Single Shot Detector), 2DCNN (2D Convolutional Neural Network), 3DCNN (3D Convolutional Neural Network), and LSTM (Long Short-Term Memory). & Large-scale hand sign language dataset presented & Complex model. & Accuracy: 96.2\% for the RKS-PERSIANSIGN dataset, 82.10\% for First-Person datasets, and an average estimation error: 12.82 for the NYU dataset. & Video camera (RGB). & 100 sign words, 10 subjects, 10 various environment. & Persian signs language. & First-Person (45 hand action, 100 K frames), NYU (36 joints, 81,009 image sequence), and RKS-PERSIANSIGN (100 signs, 10,000 samples) datasets. \\ \hline
2020 & \citet{sincan2020autsl} & CNNs (Convolutional Neural Networks), Feature Pooling Module, unidirectional and bidirectional LSTM (Long Short-Term Memory). & New AUTSL dataset presented &
Accuracy is lagging state-of-the-art methods. It is affected by the dynamic background. & AUTSL dataset based recognition model accuracy: 95.46\%, Montalbano dataset based recognition model accuracy: 96.11\% and for user-independent benchmark dataset accuracy: 62.02\%. & Microsoft Kinect v2. & AUTSL and Montalbano datasets, 226 signs, 43 subjects. & Turkish Sign Language (TSL). & AUTSL: 38,336 samples, Montalbano: 14,000 samples. \\ \hline
2019 & \citet{dawod2019novel} & HCRF (Hidden Conditional Random Field), and RDF (Random Decision Forest). & Higher recognition rates & The authors fail to perform quantitative-based analysis. & HCRF classifier mean accuracy (numbers): 99.99\% and alphabets: 99.9\%, and the RDF mean accuracy (number):96.3\% and (alphabets): 97.7\%. & Kinect sensor v2. & A-Z alphabets and 1-20 numbers, 30 subjects (signer). & American Sign Language. & 345000 samples. \\ \hline
2019 & \citet{fatmi2019comparing} & ANN, SVM, and HMM. & Functioning ability on a PC with Bluetooth Low-Energy (BLE) connections. & The limitation is not considered non-manual, and dictionary  size is limited. & 1. ANN: 93.79\%, 2. HMM: 85.90\%, and 3. SVM: 85.56\%. & Myo Armband (x2) (sense motion and depth). & 13 ASL gestures signs, 3 subject. & American Sign Language. & 26,000 instances captured, among 66 \% used for training and remaining used for testing. \\ \hline
2018 & \citet{jiang2018feasibility} & Linear Discriminant Analysis (LDA). & Simpler than forearm device. & The limitation is that calibration is required, and  classification accuracy depends on calibration, and time also elapsed. & Accuracy for air gesture: 92.6\% and for surface  gestures: 88.8\%. & Sensing fusion using 4-Channels sEMG Electromyography, one IMU inertial measurement unit. & 8 air gestures and 4 surface gestures with 2 distinct force levels, 10 subjects. & Human-computer interaction: Hand gesture Sign (Signing Exact English). & 3 trials of datasets. \\ \hline
2018 & \citet{mummadi2018real} & Naïve Bayes, Feed-forward Neural Network(MLP), Random Forest are used as a classifier. A complementary Filter with a coefficient  factor of 0.93 was used to obtain low drift and less noise data. & 63 milliseconds taken to recognize the sign (settling times and delays are faster and minimal respectively than local fusion algorithm with IMU motion sensor-based method). & The performance of the model is depending on sensor noise and drift. Distance between sensor and hand increase, accuracy decreased. & Naïve Bayes method accuracy: 89.1\%, and FI score: 87\%, MLP method accuracy: 92.2\%, and FI score: 91.1\%, RF method accuracy: 92.95\%, and FI score: 91.98\%. Mean accuracy: 92\% and mean F1 score: 91\%. & 3D magnetometer, 3D gyroscope (GYRO), and 3D accelerometer (ACC). & 22 hand gestures, 57 subjects. &
French Sign Language (LSF). & 1.25 million samples. \\ \hline
2018 & \citet{kim2018finger} & Ensemble Feed-forward Neural Networks. & Easy to wear, adaptable to portable devices. & Selection of hyperparameter was not addressed, suffered by convergence and computation problem. & E- ANN with 8 Classifiers accuracy: 97.4\%. & Armband module (8-channel electromyography) sensors. & 7 numbers, 17 vowels and 14 consonants, 17 subjects. & Korean finger language. & 1500 training data. \\ \hline
2018 & \citet{jane2018sign} & Artificial neural network, wavelet denoising techniques, and TKEO (TeagerKaiser energy operator). & Easy to implement. & Two-handed and user-independence signs recognition was not addressed. & Average accuracy: 97.12\%. & Myo Armband (gyroscope, accelerometer, magnetometer, and sEMG (surface electromyography)) sensors. & 48 lexicon word. & Signing Exact English (SEE-II). & 4927 samples, 204 features. \\ \hline
2018 & \citet{li2018skingest} & K-Nearest Neighbour, Linear Discriminant Analysis, and Support Vector Machines. & Less interference, stretchable and wearable comfort. & The model was not robust in nature. Hysteresis characteristics and noise present in the sensor lead to misclassification. & LDA accuracy: 97.81\%, KNN accuracy: 97.86\%, SVM accuracy: 97.89\%, and ASL recognition for 0-9 average accuracy: 98\%. & Firmly stretchable strain sensor (Custom Glove). & Six subjects. 0-9 ASL sign. & American sign language. & 10 trials of each gesture by 6 subjects, 540 data samples. \\ \hline
2018 & \citet{yin2018research} & Template Matching, BP neural network, and Combined Model. & High recognition rate data glove. & Dynamic gesture recognition based research was not performed, and data glove barrier is there, and only a single background effect considered for the experiment fail to generalize in other background. & 1. Template Matching accuracy: 96.7\%, 2. Feed-forward Neural Network accuracy: 98.4\%, combined model accuracy: 99.8\%. & Custom Glove (Bend Sensors (x5), FLEX2.2). & 6 numbers, 3 letters, 5 subjects. & Hand gesture sign. & 1000 different data for each gesture from 5 different signers, a total of 9000 data. \\ \hline
2018 & \citet{lee2018smart} & Support Vector Machines. & Custom-made devices not required because 3D-printed based device used it can fit different sign irrespective of hand and finger sizes holders. & The crucial factor like background light and other effects fail to be addressed. Sign respect to words and sentences not considered for experiments, and implementation can be on a smaller-sized printed circuit board if the research gap is not addressed. & Without pressure sensor accuracy: 65.7\%, with a fusion of pressure sensors accuracy: 98.2\%. & Custom Glove (9-Axis IMU, Flex Sens.(x5), Pressure Sens. (x2)), five flex sensors, two pressure sensors, and a three-axis inertial motion sensor. & 28 gesture patterns (26 ASL letters, 2 signs), 12 subject. & American sign language. & 6,480,000 datasets (12 subjects× 20 times × 10 s × 100 Hz × 27 signs). \\ \hline
2017 & \citet{yang2017chinese} & Multi-Stream Hidden Markov Models, and Decision Tree classifier. & Time consumption is reduced, and recognition accuracy is improved by the decision tree based classifier. & The proposed model result still lagging behind the literature state of the methods. & User dependent model accuracy: 94.31\%, user-independent model accuracy: 87.02\%. & 4-Channel sEMG, 3-Axis (gyroscope, accelerometer) sensor. & 150 signs (one-handed sub-words sign: 81,  two-handed sub-words sign: 69, hand orientations: 3, hand amplitude levels: 3), 8 subjects. & Chinese Sign Language (CSL). & 30000 sub word data sample (total 3750 sub-word samples per subject). \\ \hline
2016 & \citet{wei2016component} & Code matching method, and fuzzy K-mean algorithm. & User’s training burden minimized. & Target set size is limited; instead of code matching method, advanced fusion method can improve the recognition accuracy. & The recognition accuracy of two reference subject for one-third gestures of the target set: (82.6 ± 13.2)\%, and (79.7 ± 13.4)\% and half of the target set: (88 ± 13.7)\% and (86.3 ± 13.7)\%. & sEMG (surface electromyographic), GYRO (gyroscopes), and ACC (accelerometers). & 110 sign words, 5 subjects. & Chinese Sign Language (CSL). & 13750 (2750 sign word samples for each subject). \\ \hline
2016 & \citet{chansri2016hand} & Histograms of oriented gradients, and artificial neural network (BP). & Simple model. & The author can not perform statistical analysis. & Accuracy: 84.05\%. & Microsoft Kinect (color and depth). & 42 letters, 24 hand gestures. & Thai Sign Language. & 420 hand gesture samples. \\ \hline
2014 & \citet{almeida2014feature} & Support Vector Machines (SVM). & Simplified model with generic nature. & Fail to address the aspect of feature selection and recognition rate based on uncertainty presented in the suggested model. & Average accuracy: 80\%. & Kinect sensor, nuiCapture Analyze software (RGB-D sensor). & 34 specific sign, 1 subject. & Brazilian Sign Language. & 170 video samples. \\ \hline
2012 & \citet{ong2012sign} & Sequential Pattern Trees Boosting algorithm. & Run-time complexity is less. & Stability issue. & Accuracy: 55\% for the first ranked sign and 87\% for within the top 10 signs. & Kinect \textsuperscript{TM} camera. & 1. 40 signs, 14 subjects, 2. 982 signs, 1 subject. & DGS (German Sign Language (Deutsche Gebärdensprache), Greek Sign Language (GSL). & 2800 samples for DGS Kinect 40 dataset and 4910 samples for GSL 982 Signs dataset. \\ \hline
2009 & \citet{hruz2009input} & Hidden Markov Model, and kiosk. & Less difficult concern to usability. & Dataset is tiny, and performance improvement needs more data. & 8 states HMM Model recognition rate: 81.63\%. & Camera. & 50 sign, 2 subject. & CzSL (Czech sign language). & 338 samples.\\ \hline
\end{longtable}
\end{landscape}
\twocolumn
It makes SLR performance way behind performance of speech recognition. The existing research work on continuous manual SLR are as follows:
\par \textbf{Traditional methods:} \citet{nayak2012finding} pointed out the feature extraction approach for continuous sign. Relational distribution are captured from the face and hand present in the images. The parameters are optimized by ICM, so convergence speeds up; they used dynamic time warping for distance computation between two sub-strings. The continuous sign sentence extracts the recurrent features using RD, DWT, and ICM based approaches. \citet{kong2014towards} performed continuous SLR by merging of CRF (conditional random field) and SVM in a framework of Bayesian network. They performed a semi-Markov CRF decoding scheme-based merge approach for independent continuous SLR. 
\citet{tripathi2015continuous} carried out a gesture recognition model for continuous Indian Sign Language. They extracted meaningful gesture frames using the Key-frame extraction method. The orientation histogram technique extracted each gesture-relevant feature and used the Principal Component Analysis to reduce the feature dimension. They used the distance classifier for classification. According to performance analysis with other considered classifiers, the Correlation and Euclidean distance-based classifier perform with a better recognition rate. \citet{gurbuz2021american} developed an ASL model for the RF sensing-based feature fusion approach. They use LDA, SVM, KNN, and RF as classifiers. The random forest classifier-based model for five signs results in 95\% recognition accuracy, while 20 signs result in 72 \%. They can use the deep learning classifier in the future to improve recognition accuracy. \citet{hassan2019multiple} proposed Modified k-Nearest Neighbor and Hidden Markov Models based on Continuous Arabic SLR. Window-based statistical features and 2D DCT transformation extract the features. The proposed model performance analyzed with sensors, vision-based datasets, and motion tracker dataset leads to a better recognition rate. For sentence recognition (MKNN) Modified k-Nearest Neighbor yields the best recognition rate than the HMM-based Toolkit. For word recognition, RASR performs better with a higher recognition rate than MKNN GT2K.
\par
\begin {figure}
\includegraphics [scale= 0.5]{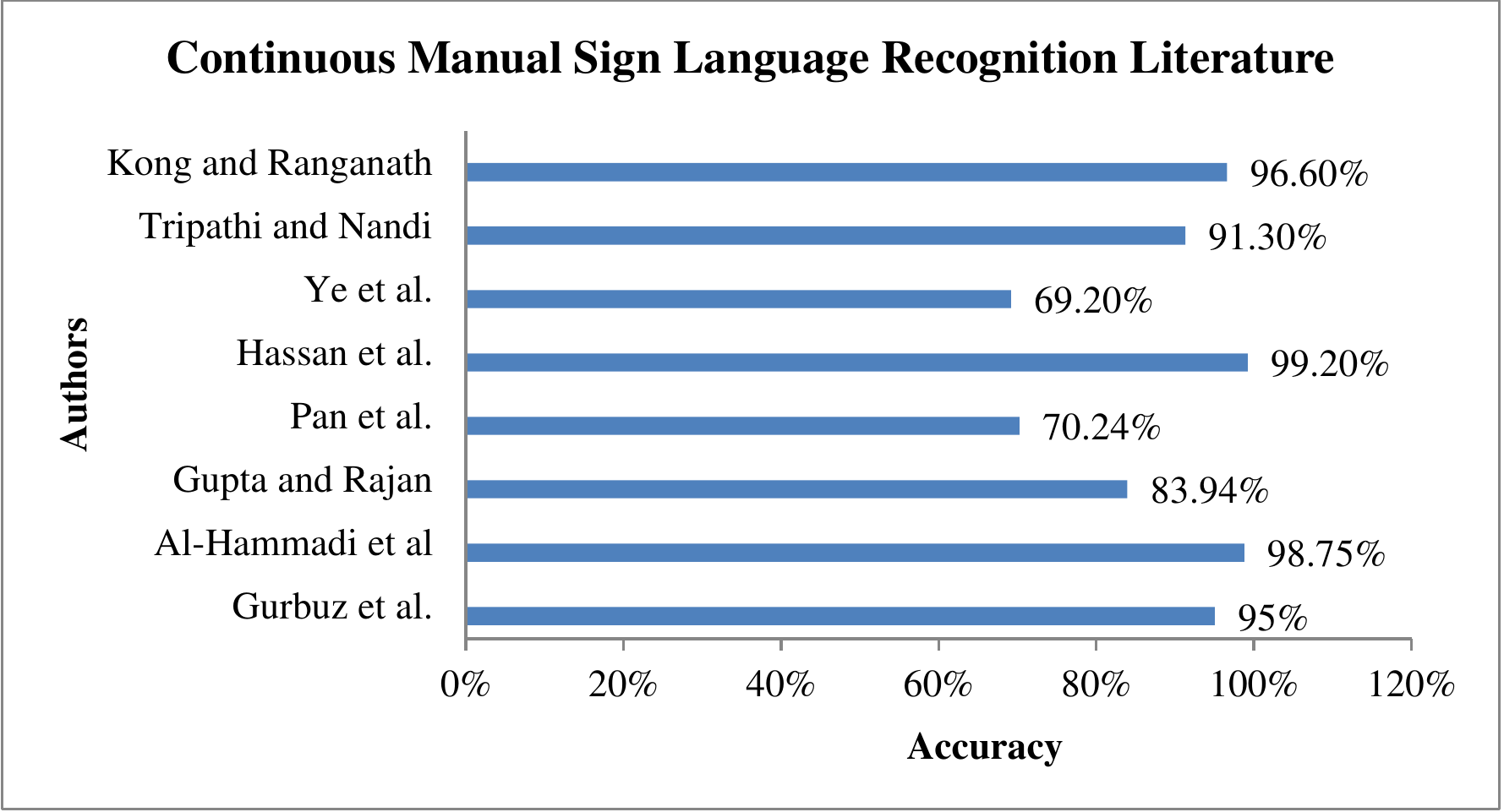}
\caption{Continuous Manual SLR Literature. This bar chart presents the mean accuracy and recognition rate achieved in the literature on continuous manual SLR concerning different sign languages, models, and modalities.}
\centering
\label{Figure 7}
\end{figure}

 \textbf{CNN, LSTM and Cross model based related work on continuous manual SLR:}
\citet{ye2018recognizing} pointed out a 3D convolutional neural network (3DCNN) with a fully connected recurrent neural network (FC-RNN) to localize the continuous video temporal boundaries and recognize sign actions using an SVM classifier. Designed Convolutional 3D and recurrent neural network-based integrated SLR model for continuous ASL sign recognition. \citet{al2020deep} presented a single modality-based feature fusion adopted 3DCNN model for dynamic hand gestures recognition. They captured the hand feature using an open pose framework. MLP and auto encoder-based feature extracted 3DCNN model with open pose framework based on hand sign capturing model result in good recognition accuracy for KSU-SSL (King Saud University Saudi Sign Language) dataset using a batch size of 16. \citet{gupta2020comparative} examined the performance of three models, namely modified time-LeNet, t-LeNet (time-LeNet), and MC-DCNN based on Indian SLR. continuous Indian SLR models based on MCDCNN, t-Lenet, and modified t- Lenet classifier using sensor-based dataset presents and performance-based investigation carried out. \citet{pan2020attention} spatial and temporal fused Attention incurred Bi directional long term memory network-based SLR model developed. They detected captured video key action by optimKCC. Multi-Plane Vector Relation (MPVR) is used to get skeletal features. They performed two dataset-based analyses to prove the validity of continuous Chinese SLR concerns sign independent and dependent cases. \citet{papastratis2020continuous} suggested a cross-modal learning-based continuous SLR model, and they have proved validity with three public datasets, namely RWTH-Phoenix-Weather-2014, RWTH- Phoenix-Weather-2014T, and CSL. They achieved the performance improvement of the suggested model by considering additional modalities. \\ Table \ref{Table 5} and Figure \ref{Figure 7} provide a better understanding of literature work regarding continuous manual SLR.
\begin {figure}
\includegraphics[scale= 0.9]{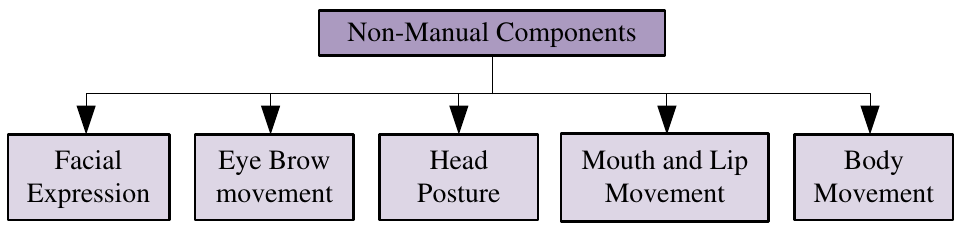}
\caption{Non-Manual Components. We showed the important non-manual features related to sign language. These non-manual components are helpful to make sensible recognition of similar types of signs.}
\centering
\label{Figure 8}
\end{figure}
\subsection{Non-Manual SLR} Facial expressions, head movement, mouth movement, eye movement, eyebrow movement, and body posture are the non-manual sign parameters. Non-manual sign components showed in Figure \ref{Figure 8}. A facial expression considering the lowering and raising of eyebrows expresses grammatical information and emotions. Signers are good listeners and follow eye contact. Similar hand pose signs can be recognized by considering non-manual features. The isolated and continuous are the two types of non-manual SLR models. 
\subsubsection{Isolated Non-Manual SLR} The study of related research works in isolated non-manual-based SLR as follows:
\par \textbf{HMM based work:} \citet{von2008recent} designed Hidden Markov Model-based British SLR with manual and non-manual features. \citet{aran2009signtutor} performed a Turkish SLR model using a cluster-based Hidden Markov Model. They proved the validity by cross-validation with eight folds (sign independent) and five folds (sign dependent). \citet{sarkar2011segmentation} presented an isolated American SLR model using Hidden Markov Model. They improved the segmentation process by a dynamic programming-based approach. \citet{fagiani2015signer} carried out the Hidden Markov Model-based isolated Italian sign recognition in concern to signer independence. The suggested model gets better accuracy than the support vector machine-based recognition model. \citet{zhang2016chinese} suggested adaptive hidden states incurred Hidden Markov Model for Chinese SLR. The carried-out fusion of trajectories and hand shapes leads to better recognition. \citet{kumar2018independent} performed an Indian SLR model based on a decision fusion approach with two modalities (facial expression and hand gesture). They used an HMM-based classifier for recognition and used IBCC for decision fusion purposes. They have carried out two modalities (facial expression and hand gesture) associated with IBCC based on HMM classifier decision fusion approach for Indian SLR. Using advanced classifiers and feature extraction algorithms can improve recognition accuracy.
\par
\begin {figure}
\includegraphics[scale= 0.36]{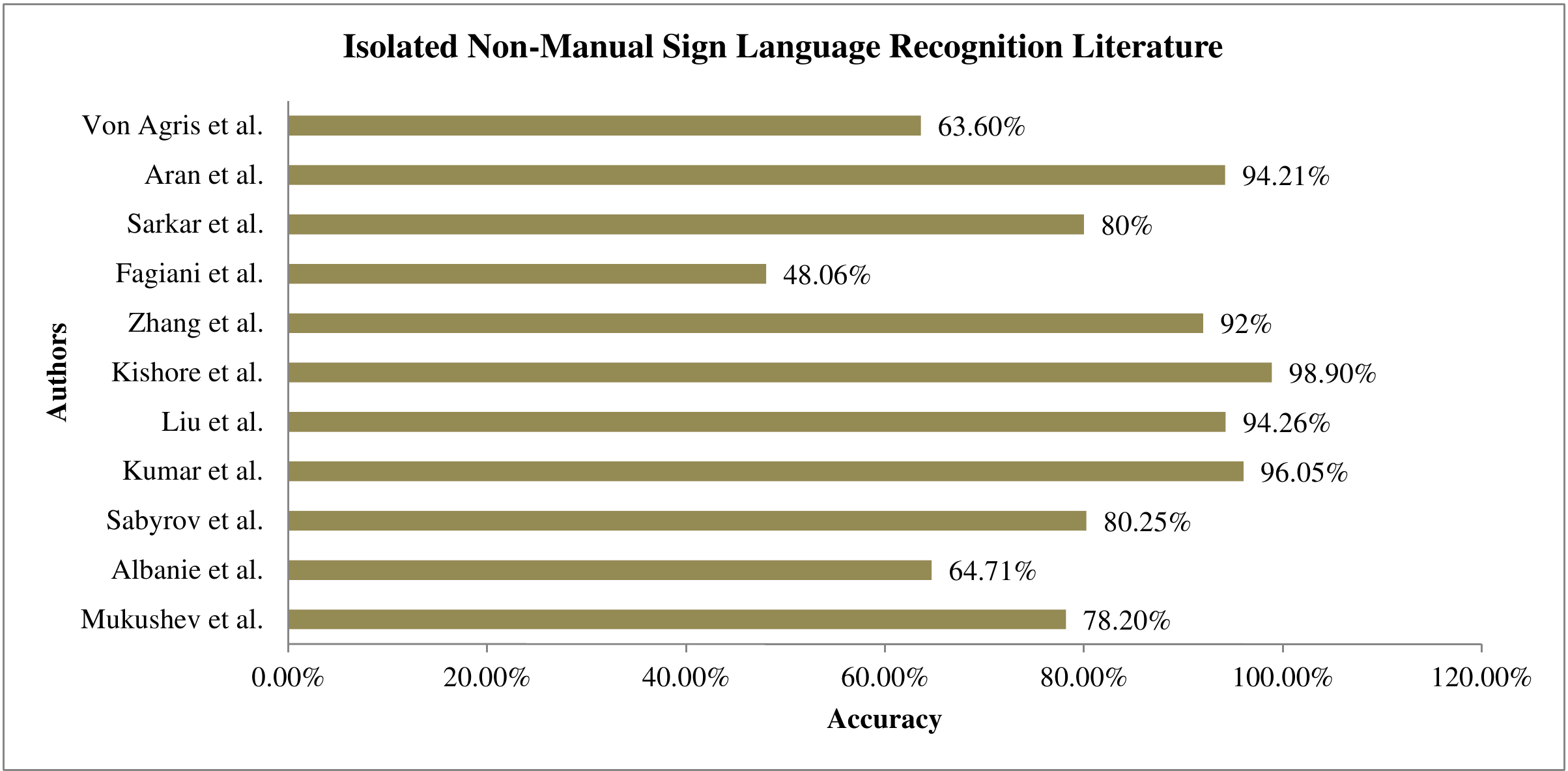}
\caption{Isolated Non-Manual SLR Literature. This bar chart represents the mean accuracy and recognition rate achieved in the literature concerning the isolated non-manual SLR regarding the different sign languages, models, and modalities.}
\centering
\label{Figure 9}
\end{figure}

 \textbf{Logistic regression and CNN based work:} \citet{sabyrov2019towards} developed K-RSL(Kazakh-Russian Sign Language) interpreted as a human-robot model using Logistic Regression with incurred non-manual components. \citet{mukushev2020evaluation} performed Logistic Regression-based SLR using manual and non-manual features. In the captured video, they got key points using OpenPose. \citet{kishore2018motionlets} performed Adaptive Kernels Matching algorithm that incurred 3-D Indian SLR model claims improved classification accuracy compared with state-of-the-art methods. Better classification accuracy achieved by 3D motion capture models than Microsoft Kinect and leap motion sensor-based model.\citet{liu2018self} pointed out ST-Net (Spatial-Temporal Net) associated with self-boosted intelligent systems for Hong Kong SLR. Compared to a Kinect-based system, the suggested approach performs well with a better recognition rate. \citet{albanie2020bsl} proposed a Spatio-temporal convolutional neural network-based British SLR model. The pretraining has improved by presenting new larger-scale data, namely BSL-1K.
\\ We perform a comprehensive study of the recent developments concerning non-manual SLR. Table \ref{Table 6} and Figure \ref{Figure 9} show isolated non-manual SLR-related literature work to make a clear understanding.\\
\subsubsection{Continuous Non-Manual SLR}
Continuous non-manual SLR is highly complex because the issue related to the context sequence has to be handled appropriately for effective performance or enriched accuracy \cite{wilbur2013phonological}. The temporal boundaries-related problem makes continuous SLR a complex and arduous task. We discuss the related research work as follows:
\par \textbf{Classical methods:} \citet{farhadi2006aligning} carried out the HMM-based continuous ASL to English subtitles alignment model. With simple HMMs based on a discriminative word model, they perform word spotting. \citet{infantino2007framework} developed a common-sense engine integrated self-organizing map (SOM) neural network-based SLR model for LIS (Italian sign language). \citet{sarkar2011segmentation} performed a HMM-based continuous ASL. They used a dynamic programming-based approach to improve the segmentation. \citet{forster2013modality} pointed out the German SLR model using Multi-stream HMMs based on combination methods. Compared to system combination and feature combination approaches, synchronous and asynchronous combination-based models achieved better performance. \citet{yang2013robust} presented CRF and SVM associated with a continuous ASL using both manual and non-manual features. BoostMap embeddings verified the hand shape, segmenting done by hierarchical CRF, and recognition was performed using SVM. \citet{zhang2016multi} suggested a Linear SVM based on an automatic ASL by fusing five modalities. The large-scale dataset-based investigation could be future work to improve recognition accuracy.\\

\clearpage
\onecolumn
\begin{landscape}
\tiny
\begin{longtable}[c]{|p {0.4 cm}|p{1 cm}| p {2.5 cm}|p {2.5 cm}|p{2.5 cm}|p {3.5 cm}| p{1.5 cm}|p {2 cm}|p{1.7 cm}|p{2 cm}|}
\caption{Continuous Manual SLR Literature Work. Related work concerning the vision and sensor-based SLR model concerns the continuous manual sign, comprehensively summarized here.}
\label{Table 5}\\
\hline
\textbf{Year} &
\textbf{Author} &
\textbf{Method} &
\textbf{Pros} &
\textbf{Cons} &
\textbf{Accuracy or Result} &
\textbf{Sensor} &
\textbf{Sample and Lexicon Size} &
\textbf{Model Type} &
\textbf{Dataset}\\
\hline
\endfirsthead

\hline
\textbf{Year} &
\textbf{Author} &
\textbf{Method} &
\textbf{Pros} &
\textbf{Cons} &
\textbf{Accuracy or Result} &
\textbf{Sensor} &
\textbf{Sample and Lexicon Size} &
\textbf{Model Type} &
\textbf{Dataset}\\
\hline
\endhead
\hline
\endfoot
\endlastfoot
\hline
2021 & \citet{gurbuz2021american} & Principal Component Analysis (PCA), short term Fourier transform, and RBF (Radial Basis Function) associated SVM (Support Vector Machine). & Contactless sensing, environment independent & The author can not perform a comparative analysis with the existing method. & Recognition accuracy for 20 sign, 150 features: 72.5\% and for 5 sign: 95\%. & RF sensors and Kinect sensor. & 20 signs, 7 subjects. & American SLR. & 240 samples. \\ \hline
2020 & \citet{al2020deep} & 3D convolutional neural network (3DCNN), auto-encoders, Multilayer perceptron, and open pose framework. & The expense concerning the training is minimal. & Model results in good accuracy for a small batch. & MLP based feature extraction model achieved recognition accuracy as 98.62\% for sign dependent and 87.69\% for sign independent and auto encoder based feature extraction model achieved recognition accuracy as 98.75\% for sign independent and 84.89\% for sign independent. & RGB video camera. & 40 dynamic hand gestures, 40 subjects. & Saudi Sign Language. & 8000 data samples. \\ \hline
2020 & \citet{gupta2020comparative} & Modified time-LeNet, MC-DCN (Multi channel deep convolution neural network), t-LeNet (time-LeNet), and SGD (stochastic gradient descent). & Sensor-based on continuous sign recognition model. & Trainable parameters are a large, over-fitting problem. & MCDCNN accuracy: 83.94\%, modified t-LeNet accuracy: 81.62\%, and t-LeNet accuracy: 79.70\%. & Wireless IMUs (inertial measurement units). & 11 sentences with 15 words, 10 subjects. & Indian Sign Language. & 1100 samples. \\ \hline
2020 & \citet{papastratis2020continuous} & Cross-modal learning approach. & Recognition of accuracy improved by latent representations. & Sensitive to Multi modalities. & 1. WER: 24.0\% (RWTH-Phoe-nix-Weather-2014), 2. WER: 24.3\% (RWTH- Phoe-nix-Weather-2014T), 3. WER: 2.4\% (CSL). & Stationary color camera, Kinect 2.0. & For Dataset 1: 9 subjects, vocabulary: 1088,dataset 2: 9 subjects, sign:1295, dataset 3: 50 subject,  100 sentences. & German sign language, Chinese Sign Language & RWTH-Phoenix-Weather-2014, RWTH- Phoenix-Weather-2014T, CSL. \\ \hline
2020 & \citet{pan2020attention} & Attention-Based BLSTM, KCC (key-frame-centered clips), and MPVR (Multi-Plane Vector Relation). & Speed up convergence. & Real-time practical implementation was lacking. & Fused Attention incurred BLSTM with optimum KCC and MVPR based sign recognition model accuracy: 70.24\% (CSL dataset) and 60.31\% (DEVISIGN dataset) for sign dependent and 68.32\% (CSL dataset) and 58\% (DEVISIGN dataset) for sign independent. & Kinect-1.0, Kinect-2.0. & DEVISIGN dataset: 500 sign words, 8 subjects, CSL dataset: 10 subjects, 200 sign words. & Chinese Sign Language. & CSL dataset: 20000 samples. \\ \hline
2019 & \citet{hassan2019multiple} & Modified k-Nearest Neighbor, and Hidden Markov Models (GT\textsuperscript{2} K and RASR). & Require less computation time. & Robustness needs to be improved. & For sentence recognition, MKNN based model: 97.78\% (DG5-VHand dataset), and for words recognition RASR based model: 99.20\% (DG5-VHand dataset). & Camera, DG5-VHand data glove and Polhemus G4 motion tracker. & 80 sign, 40 sentences, 1 subject. & ArSL (Arabic sign language). & 800 words samples, 400 sentence samples. \\ \hline
2018 & \citet{ye2018recognizing} & 3DRCNN (3D recurrent convolutional neural networks) with SVM classifier. & RGB, motion, and depth channels fusing lead to better accuracy. & Facial information was not considered. Hence, the performance is poor. & Accuracy: 69.2\% for Person-dependent and an accuracy: 65.8\% for Person-independent. & Kinect 2.0 sensor. & 99 sign, 5 sentences, 15 subjects. & American Sign Language. & Sentence videos (100) and Sequence videos (27) are collected. \\ \hline
2015 & \citet{tripathi2015continuous} & OH (Orientation Histogram), PCA (Principal Component Analysis), and distance-based classifier. & New ISL continuous dataset presented. & Misclassification for a similar type of gesture. & Average recognition rate for Euclidean distance: 90\% (18 bins OH), 91.3\% (36 bins OH), and Average recognition rate for correlation: 89\% (18 bins OH), 89.8\% (36 bins OH). & Canon EOS camera. & 10 sentences, 5 subjects & Indian Sign Language. & 500 video samples. \\ \hline
2014 & \citet{kong2014towards} & Semi-Markov CRF decoding scheme based probabilistic approach. & Complexity related to decoding reduced. & Larger vocabularies' scalability issue. & For an unseen sign from considered signer, accuracy: 96.6\% and recall rate: 95.7\%, for unseen signers' accuracy: 89.9\% and recall rate: 86.6\%. & CyberGlove(x2) and 3D Trackers. & 74 sentences (107 ASL signs), 8 subjects. & American sign language. & 2393 sentences and 10852 sign instances. \\ \hline
2012 & \citet{nayak2012finding} & Relational Distributions, ICM (iterative conditional modes) algorithm, and DTW (Dynamic Time Warping). & Aid for faster training set generation. & They did not address the amplitude variation of the various signer. Comparative qualitative and quantitative analysis are lacking. & NA & Video camera. & 155 American Sign Language (ASL), 12 groups. & American Sign Language. & 155 video sequences. \\ \hline
\end{longtable}
\end{landscape}
\clearpage
\begin{landscape}
\tiny
\begin{longtable}[c]{|p {0.4 cm}|p{1 cm}| p {2.5 cm}|p {2.5 cm}|p{2.5 cm}|p {3.5 cm}| p{1.5 cm}|p {2 cm}|p{1.7 cm}|p{2 cm}|}
\caption{Isolated Non-Manual SLR Literature Work. Related work with regard to vision and sensor based SLR model concern the isolated non-manual sign comprehensively summarized in the tabular form for better understanding.}
\label{Table 6}\\
\hline
\textbf{Year} &
\textbf{Author} &
\textbf{Method} &
\textbf{Pros} &
\textbf{Cons} &
\textbf{Accuracy or Result} &
\textbf{Sensor} &
\textbf{Sample and Lexicon Size} &
\textbf{Model Type} &
\textbf{Dataset}\\
\hline
\endfirsthead
\hline
\textbf{Year} &
\textbf{Author} &
\textbf{Method} &
\textbf{Pros} &
\textbf{Cons} &
\textbf{Accuracy or Result} &
\textbf{Sensor} &
\textbf{Sample and Lexicon Size} &
\textbf{Model Type} &
\textbf{Dataset}\\
\hline
\endhead
\hline
\endfoot
\endlastfoot
\hline
2020 & \citet{mukushev2020evaluation} & Logistic Regression. & Non-manual components considered as input lead to better accuracy. & Need improvement concern to accuracy. & Accuracy: 78.2\%. & LOGITECH-C920 HD PRO WEBCAM. & 20 signs, 5 subjects. & K-RSL (Kazakh-Russian sign language). & 5200 samples. \\ \hline
2020 & \citet{albanie2020bsl} & Spatio-temporal convolutional neural network. & Large-scale dataset (BSL-1K) presented. & Fails to address visual similarity issue. & Accuracy: 46.82\% (WLASL) and 64.71\% (MSASL) top-1 case. & Camera. & MS-ASL: 1000 words, 222, WSASL:2000 words, 119 subjects BSL-1K: 1064, 40subjects. & BSL (British Sign Language), ASL (American Sign Language). & WSASL: 21K MS-ASL: 25K BSL-1K: 273K. \\ \hline
2019 & \citet{sabyrov2019towards} & Logistic Regression. & Huge dataset not required a humanoid robot sign interpreter. & Experimentation in real-world is lacking. & For 20 sign dataset accuracy: 73\% and for 2 class accuracy: 80.25\%. & LOGITECH C920HD PRO WEBCAM. & 20 words, 3 subjects. & K-RSL n (Kazakh-Russian Sign Language). & 2000 videos. \\ \hline
2018 & \citet{kumar2018independent} & HMM (Hidden Markov Model) and IBCC (Independent Bayesian Classification Combination). & The suggested model performed better than BLSTM-NN. & Possibility to classify wrongly fails to present a comparative analysis based on recent methods for the considered applications. & Double hand gestures based recognition rate: 94.27\%, and the single-hand sign based recognition rate:  96.05\%. & Leap Motion and Kinect sensor. & 51 dynamic sign words (31 sign with two hands, 20 sign with a single hand), 10 subjects. & Indian Sign Language (ISL). & 4080 data samples include both single and two hands sign. \\ \hline
2018 & \citet{liu2018self} & ST-Net (SpatialTemporal Net). & Light-weight and robust. & Require more computation time. & Accuracy: 94.26\% for Person dependent and 91.19\% for Person independent sign. & Microsoft  Kinect. & 227 words, (86 words – single hand, 33 words), 2 hands separated, and 108 words- 2 hands intersected, 3 subjects based 1802 mouth images. & HKSL (Hong Kong Sign Language). & 5221 samples videos. \\ \hline
2018 & \citet{kishore2018motionlets} & Adaptive Kernels Matching algorithm. & Overcome the issue of spatio-temporal misalignment, Small action changes discovered effectively. & The author fails to perform real-time, live data-based analysis. & Accuracy: 98.9\% & Kinect and leap motion sensors. & Five sets of data with  500 signs, 5 subjects each set. & IndianSL (Indian Sign Language). & 18000 signs with 36 variations per sign comprises a testing set. \\ \hline
2016 & \citet{zhang2016chinese} & Adaptive Hidden Markov Models, enhanced shape context. & Computational cost reduced. & Fail to assure accuracy for the larger dataset. & Accuracy rate for dataset 1: 92\% in the top 1, 99\% in the top 5, and 100\% in the top 10. For dataset 2 accuracy: 86\% in the top 1, 96.8\% in the top 5, and 98.8\% in the top 10. & Microsoft Kinect. & 1. 100 signs 1 subject, 2. 500 signs 1 subjects, 99 signs, 5 subjects. & CSL (Chinese sign language). & 1. 500 videos, 2. 2500 videos. \\ \hline
2015 & \citet{fagiani2015signer} & Hidden Markov Models. & Easy to implement in real-time case. & Overlapping issue. & Average accuracy rate: 48.06\%. & Digital Video Camera. & 147 signs, 10 subjects. & LIS (Italian sign language / Lingua Italiana dei segni). & 1,470 videos samples (A3LIS-147). \\ \hline
2011 & \citet{sarkar2011segmentation} & Dynamic programming methods, and Hidden Markov Models. & Adaptable to uncontrolled domain. & Sensitive to dataset. & Recognition rate: 80\%. & Video Camera. & 147 signs, 10 subjects. & ASL (American sign language). & 294 samples. \\ \hline
2009 & \citet{aran2009signtutor} & Hidden Markov Models with cluster algorithm. & Classify the similar sign accurately. & The small dataset used for validation. & Average accuracy rate: 79.1\% (Signer-independent) and 94.21\% (Signer-dependent). & Camera. & 19 signs,  8 subjects. & TSL (Turkish sign language). & 760 samples for each fold (8 folds for Signer-independent and 5folds for Signer-dependent). \\ \hline
2008 & \citet{von2008recent} & Hidden Markov Models. & User-friendliness. & Recognition accuracy decay with low resolutions. & Average recognition rates: 63.6\%. & Video camera & 263 signs, 4 subjects. & BSL (British sign language). & 8100 video samples. \\ \hline
\end{longtable}
\end{landscape}
\clearpage
\onecolumn
\begin{landscape}
\tiny
\begin{longtable}[c]{|p {0.4 cm}|p{1 cm}| p {2.5 cm}|p {2.5 cm}|p{2.5 cm}|p {3.5 cm}| p{1.5 cm}|p {1.5 cm}|p{1.8 cm}|p{2.5 cm}|}
\caption{Continuous Non-Manual SLR Literature Work. Related work concerning the vision and sensor-based SLR model concern the continuous non-manual sign, comprehensively summarized here.}
\label{Table 7}\\
\hline
\textbf{Year} &
\textbf{Author} &
\textbf{Method} &
\textbf{Pros} &
\textbf{Cons} &
\textbf{Accuracy or Result} &
\textbf{Sensor} &
\textbf{Sample and Lexicon Size} &
\textbf{Model Type} &
\textbf{Dataset}\\
\hline
\endfirsthead
\hline
\textbf{Year} &
\textbf{Author} &
\textbf{Method} &
\textbf{Pros} &
\textbf{Cons} &
\textbf{Accuracy or Result} &
\textbf{Sensor} &
\textbf{Sample and Lexicon Size} &
\textbf{Model Type} &
\textbf{Dataset}\\
\hline
\endhead
\hline
\endfoot
\endlastfoot
\hline
2020 & \citet{brock2020recognition} & Convolutional Neural Network. & Complex linguistic content handling ability. & Expensive. & Average Word Error Rate (WER): 15.71\% for Non-Manual Expression label. & Single-camera video. & 1432 signs, 1 subject. & JSL (Japanese Sign Language). & DJSLC corpus a total of 1432 sequential sign. \\ \hline
2020 & \citet{zhou2020spatial} & Spatial-temporal multi-cue (STMC) network, spatial multi-cue (SMC) module and a temporal multi-cue (TMC) module, joint optimization strategy. & Able to handle different cues at the same time. & Training time and complexity is high. & STMC WER: 28.6\% (CSLunseen), 20.7\% (PHOENIX-2014), 21.0\% (PHOENIX-2014-T). & Video camera. & PHOENIX-2014: 9 subjects 1295 sign, PHOENIX-2014-T 1115 for sign gloss CSL: 10 subjects, 500 words. & CSL (Chinese sign language), DGS (German Sign Language). & CSL:5000 videos, PHOENIX-2014:6841 samples, PHOENIX-2014-T:8257 samples.\\ \hline
2020 & \citet{koller2020weakly} & Hybrid CNN-LSTM-HMMs. & Speedup convergence. & Sensitive to overlapping and computation is difficult. & For RWTHPHOENIX-Weather 2014 dataset, WER: 26.0\%. & Video camera. & 1. 1066 signs 9 subjects, 2. 1080 sign 9 subjects. & DGS (German Sign Language (Deutsche Gebärdensprache). & One million hand shape images from 23 subjects, Phoenix14T,Phoenix14. \\ \hline
2016 & \citet{koller2016deep} & Iterative Expectation Maximization incurred CNN. & Training is easy. & Convergence took much time. & Recognition accuracy: 62.8\%. & Video camera. & 1.1080 sign 9 subjects, 2. 455 sign 1 subjects. & DGS (German Sign Language (Deutsche Gebärdensprache). & One million hand shape images Phoenix14, Signum. \\ \hline
2016 & \citet{zhang2016multi} & Linear Support Vector Machine. & Simple concatenation approach. & Unbalancing of data leads to misclassification. & Recognition rate: 36.07\% for all lexical items. & Kinect sensor. & 99 signs, 5 subjects. & ASL (American sign language). & 61 video sequences (segmented in to set of 673 video clips). \\ \hline
2013 & \citet{yang2013robust} & Conditional random field support vector machine and BoostMap embedding method. & Issues of label bias not occurred. & The author can not perform experimentation with real-time live data. & Recognition rate: 84.1\%. & Cameras. & 24 signs, alphabets-17, and facial expressions-5. & ASL (American sign language) & 98 ASL signed sentences. \\ \hline
2013 & \citet{forster2013modality} & Multi-stream HMMs based on combination methods & Robust model. & An asynchronous combination, not flexible to handle the large-scale dataset. & For SIGNUM dataset WER: 10.7\%, and for PHOENIX dataset WER: 41.9\%. & Video Camera. & 1. 266 signs 1 subjects, 2. 455 signs 1 subjects, 3. 455 signs, 25 subjects. & DGS (German Sign Language (Deutsche Gebärdensprache). & Signum dataset consists 530850 total number of frame and PHOENIX dataset consists 53033 total number of frame. \\ \hline
2011 & \citet{sarkar2011segmentation} & Hidden Markov Model. & Adaptable to uncontrolled domain. & Sensitive to dataset. & Correct detection: 92\%. & Video Camera. & 65 signs subjects. & ASL (American sign language). & 30 sentences. \\ \hline
2007 & \citet{infantino2007framework} & Self-organizing map (SOM) neural network. & Simple and robust. & Selection of input weights is difficult. & Correctly recognized sentence accuracy: 82.3\% for Exp 1(30 videos of sentences with 20 signs) and 82.5\% for Exp 2(80 videos of sentences with 40 signs). & Video camera. & 40 signs. & LIS (Italian sign language / Lingua Italiana dei segni). & 160 video samples. \\ \hline
2006 & \citet{farhadi2006aligning} & Hidden Markov Model. & The simple discriminative word model. &
The author cannot perform quantitative and qualitative results analysis. & NA & Video camera. & 31 words signs. & ASL (American sign language). & 80000 frames of film. \\ \hline
\end{longtable}
\end{landscape}
\begin{longtable}{|p {1.5 cm}|p {10.4 cm}|p {0.7cm}|p {3.7cm}|}
\caption{Sensing approach based SLR. Although the sensor-based approach provides better accuracy than the vision-based approach, the sensor-based approach is not an optimal choice in real-time applications.}
\label{Table 8}
\\ \hline
\textbf{Sensing Approach} &
\textbf{Devices} &
\textbf{Year} &
\textbf{Authors} \\ \hline
\endfirsthead
\endhead
\multirow{10}{*}{Sensor} &
3-dimensional flex sensor-based data gloves, gyroscope, accelerometer, and bending sensor &
2021 &
\citet{yuan2020hand} \\ \cline{2-4} 
&
EMG sensor &
2021 &
\citet{botros14electromyography} \\ \cline{2-4} 
&
Surface electro-myogram and inertial measurement units &
2020 &
\citet{gupta2020indian} \\ \cline{2-4} 
& 4-Channels sEMG-Electromyography, one IMU inertial measurement unit &
2018 &
\citet{jiang2018feasibility} \\ \cline{2-4} 
&
3D magnetometer, 3D gyroscope (GYRO), and 3D accelerometer (ACC) &
2018 &
\citet{mummadi2018real} \\ \cline{2-4} 
&
Armband module (8-channel electromyography) sensors. &
2018 &
\citet{kim2018finger}\\ \cline{2-4} 
&
Myo Armband (gyroscope, accelerometer, magnetometer,  and sEMG (surface electromyography)) sensors &
2018 &
\citet{jane2018sign} \\ \cline{2-4} 
&
Firmly stretchable strain sensor (Custom Glove) &
2018 &
\citet{li2018skingest} \\ \cline{2-4} 
&
Wireless IMUs (inertial measurement units) &
2020 &
\citet{gupta2020comparative} \\ \cline{2-4} 
&
CyberGlove (x2) and  3D Trackers &
2014 &
\citet{kong2014towards} \\ \hline
\multirow{11}{*}{Vision} &
Microsoft Kinect v2 &
2020 &
\citet{sincan2020autsl} \\ \cline{2-4} 
&
Video camera & 2020 &
\citet{aly2020deeparslr}\\ \cline{2-4} 
&
Video camera &
2020 &
\citet{rastgoo2020hand} \\ \cline{2-4} 
& RGB video camera &
2020 &
\citet{al2020deep} \\ \cline{2-4} 
&
Laptop camera &
2020 &
\citet{hoang2020hgm}\\ \cline{2-4} 
&
Single camera video &
2020 &
\citet{brock2020recognition} \\ \cline{2-4} 
&
\begin{tabular}[c]{@{}l@{}}LOGITECH C920 HD PRO WEBCAM\end{tabular} &
2019 &
\citet{sabyrov2019towards} \\ \cline{2-4} 
&
\begin{tabular}[c]{@{}l@{}}Kinect and leap motion sensors\end{tabular} &
2018 &
\citet{kishore2018motionlets} \\ \cline{2-4} 
&
Leap Motion and Kinect sensor &
2018 &
\citet{kumar2018independent} \\ \cline{2-4} 
&
KinectTMcamera &
2012 &
\citet{ong2012sign} \\ \cline{2-4} 
&
Video camera &
2007 &
\citet{infantino2007framework}\\ \hline
\end{longtable}
\clearpage
\begin{small}
\begin{longtable}{| p{1.1 cm} | p{1.1cm} | p{1.3cm} | p{1.3cm} | p{0.9cm} | p{0.7cm} | p{1cm} | p{1cm} | p{0.78cm} | p{3.35cm} | p{0.87cm}|}
\caption{Various Datasets in SLR. (P-Publicly available, PP-Publicly available with a password, CA-Contact Author, OR-On Request, SA-Send request along with release agreement)}
\label{Table 9}
\tiny\\
\hline
\textbf{Language Level} & \textbf{Language} & \textbf{Dataset} & \textbf{Data Type} & \textbf{Subjects}& \textbf{Classes} &	\textbf{Samples} & \textbf{Country} & \textbf{Data Size} &	 \textbf{Link} &\textbf{Data Availability} \\
\hline
\endfirsthead
\hline
\textbf{Language Level} & \textbf{Language} & \textbf{Dataset} & \textbf{Data Type} & \textbf{Subjects}& \textbf{Classes} &	\textbf{Samples} & \textbf{Country} & \textbf{Data Size} &	 \textbf{Link} &\textbf{Data Availability} \\
\hline
\endhead
\hline
\endfoot
\endlastfoot
\hline
\multirow{16}{*} {Isolated}	& Indian & IIITA -ROBITA &	Videos & -	&	23	 & 605	& India &	284 MB &	\url{https://robita.iiita.ac.in/dataset.php}	& SA\\\cline{3-11} 
\multirow{16}{*} {}& \multirow{6}{*} {} & INCLUDE \cite{sridhar2020include}	& Videos &	7 & 266 &	4287 &	India &	56.8 GB & \url {https://zenodo.org/record/4010759#.YdqY9lRfg5k} &	P \\\cline{2-11}
\multirow{16}{*} {}& \multirow{6}{*} {American} &	Boston ASL LVD &	Videos, multiple angles	& 6	 & 3300+ &	9800 &	United States	& 1-2 GB &	\url{http://www.bu.edu/asllrp/av/dai-asllvd.html}	& P \\\cline{3-11}
\multirow{16}{*} {}& \multirow{6}{*} {} & ASLLVD	& Videos (multiple angles) &	6 &	3,300 &	9,800 &	United States &	1-2 GB of each & \url {http://www.bu.edu/asllrp/av/dai-asllvd.html} &	P \\\cline{3-11}
\multirow{16}{*} {}& \multirow{6}{*} {} & Purdue RVL-SLLL & Videos	& 14 &	39	& 546 &	United States &	-	& \url{https://engineering.purdue.edu/RVL/Database/ASL/asl-database-front.htm} &	OR \\\cline{3-11}
\multirow{16}{*} {}& \multirow{6}{*} {} & ASLLVD-Skeleton &	Skeleton & -	&	3,300 & 9,800 &	United States &	1-2 GB of each &	\url{https://www.cin.ufpe.br/~cca5/asllvd-skeleton/index.html} &	P\\\cline{3-11}
\multirow{16}{*} {}& \multirow{6}{*} {} & RWTH-BOSTON-50	& Videos (multiple angles) &	3 &	50	& 483 &	United States &	295 MB &	\url{https://www-i6.informatik.rwth-aachen.de/aslr/database-rwth-boston-50.php} &	P \\\cline{3-11}
\multirow{16}{*} {}& \multirow{6}{*} {} & WLASL	& Videos &	100 &	2,000 &	21,083 &	United States &	64 GB & \url{https://dxli94.github.io/WLASL/}	 & P \\
\cline{3-11}
\multirow{16}{*} {}& \multirow{6}{*} {} & MS-ASL \cite{joze2018ms}	& Videos &	222 &	1,000 &	25,513 &	United States &	1.9 MB & \url{https://www.microsoft.com/en-us/download/details.aspx?id=100121}	 & P \\
\cline{2-11}
\multirow{16}{*} {}& \multirow{1}{*} {\makecell{Argen\\-tinian}} &	LSA64 &	Videos &	10 &	64 &	3,200	&  \makecell{Argen\\-tina}	& 1.9 GB &	\url{http://facundoq.github.io/datasets/lsa64/} &	P\\
\cline{2-11}
\multirow{16}{*} {}& \multirow{1}{*} {Chinese } &	 Isolated SLR500 	& Videos \& Depth from Kinect &	50 &	500	 & 125,000	& China	& -&	\url{http://home.ustc.edu.cn/~pjh/openresources/cslr-dataset-2015/index.html} &	PP\\
\cline{3-11}
\multirow{16}{*} {}& \multirow{2}{*} {}& NMFs-CSL &
RGB videos	& 10	& 1,067 & 32,010 &	China	 &- & \url{http://home.ustc.edu.cn/~alexhu/Sources/index.html} &	SA \\\cline{2-11}
\multirow{16}{*} {}& \multirow{1}{*}{German} &	DGS Kinect 40	& Videos (multiple angles)	& 15 &	40 &	3,000 &	Germany	& 39.7 MB & 	\url{https://www.cvssp.org/data/KinectSign/webpages/downloads.html} &	CA\\
\cline{2-11}
\multirow{16}{*} {}& \multirow{1}{*}{Greek} &	GSL isol.&	Videos \& Depth from Real Sense &	7 &	310	& 40,785 &	Greece &	155 KB &	\url{https://vcl.iti.gr/dataset/gsl/}&	P\\
\cline{2-11}
\multirow{16}{*} {}& \multirow{2}{*} {Polish} &	PSL Kinect &
Videos \& Depth from Kinect	& 1	& 30 & 300	& Poland &	~1.2 GB & \url{http://vision.kia.prz.edu.pl/dynamickinect.php} &	P\\
\cline{3-11}
\multirow{16}{*} {}& \multirow{2}{*} {}& PSL ToF &
Videos \& Depth from ToF camera	& 1	& 84 &	1,680 &	Poland &	~33 GB & \url{http://vision.kia.prz.edu.pl/dynamictof.php} &	P\\\cline{2-11}
\multirow{16}{*} {}& \multirow{1}{*}{Spanish} &	LSE-Sign &	Videos &-&		2,400 &	2,400 &	Spain &-&\url{http://lse-sign.bcbl.eu/web-busqueda/} &	CA\\
\cline{2-11}
\multirow{16}{*} {}& \multirow{1}{*}{Turkish} & BUHMAP-DB &	Videos	& 11 & 	8	& 440	& Turkey &	505 MB & \url{https://www.cmpe.boun.edu.tr/pilab/pilabfiles/databases/buhmap/} &	P\\\cline{3-11}
\multirow{16}{*} {}& \multirow{1}{*}{} & \makecell{Bosphoru\\Sign22k \\\cite{ozdemir2020bosphorussign22k}} &	Videos	& 6 & 	744	& 22,542	& Turkey &	- & \url{https://www.cmpe.boun.edu.tr/pilab/BosphorusSign/home_en.html} &	P\\\cline{3-11}
\multirow{16}{*} {}& \multirow{1}{*}{} & AUTSL &	Videos	& 43 & 226 	& 38,336	& Turkey &	- & \url{https://chalearnlap.cvc.uab.cat/dataset/40/description/} &	PP\\
\hline
\multirow{12}{*} {\makecell{Contin-\\uous} }&\multirow{1}{*}	{Indian} &	ISL-CSLTR &
Videos	 & 7 &	100 &	700	 & India &	8.29 GB & \url{https://data.mendeley.com/datasets/kcmpdxky7p/1} &	P\\\cline{3-11}
\multirow{12}{*} { }&\multirow{2}{*}{} & BVCSL3D &	Videos (RGB, depth and
skeletal) &	10 &	200 &	20,000 &	India & 256 B & \url{https://ars.els-cdn.com/content/image/1-s2.0-S1045926X18301927-mmc1.xml} &	OR\\\cline{2-11}
\multirow{12}{*} { }&\multirow{2}{*}{American} & RWTH-BOSTON-104 &
Videos (multiple angles) & 3 &	104 & 201 & United States &	685 MB & \url{https://www-i6.informatik.rwth-aachen.de/aslr/database-rwth-boston-104.php} &	P\\\cline{3-11}
\multirow{12}{*} { }&\multirow{2}{*}{} & RWTH-BOSTON-400 &	Videos &	5 &	400 &	843 &	United States & - & \url{http://www-i6.informatik.rwth-aachen.de/~dreuw/database.php} &	OR\\\cline{2-11}
\multirow{12}{*}{}& \multirow{4}{*}{Chinese}	&  Continuous SLR100	& Videos \& Depth from Kinect &	50 &	100 &	25,000	& China & - & \url{http://home.ustc.edu.cn/~pjh/openresources/cslr-dataset-2015/index.html}&	PP \\\cline{3-11}
\multirow{12}{*}{}& \multirow{4}{*}{} & {\makecell{DEVISI-\\GN-D}} & \multirow{3}{*}{Videos} & 8 & 500 &	6,000 &	China &- & \url{http://vipl.ict.ac.cn/homepage/ksl/data.html#page2} & SA\\\cline{3-11}
\multirow{12}{*}{}& \multirow{4}{*}{} & {\makecell{DEVIS-\\IGN-G}}& \multirow{3}{*}{Videos}&
8 &	36 &	432	& China	 & - & \url{http://vipl.ict.ac.cn/homepage/ksl/data.html#page2} &	SA
\\\cline{3-11}
\multirow{12}{*}{}& \multirow{4}{*}{} & {\makecell{DEVIS-\\IGN-L}} & \multirow{3}{*}{Videos} &
8 &	2000 &	24,000	& China	& - &\url{	http://vipl.ict.ac.cn/homepage/ksl/data.html#page2}& SA\\\cline{2-11}
\multirow{12}{*}{}& \multirow{3}{*}{German} &	RWTH-PHOENIX-Weather 2014 &	Videos &
9 &	1,081 &	6,841 &	Germany & 52 GB & \url{	https://www-i6.informatik.rwth-aachen.de/~koller/RWTH-PHOENIX/}&	P\\\cline{3-11}
\multirow{12}{*}{}& \multirow{3}{*}{} & SIGNUM	& Videos &	25 & 450 & 33,210 & Germany & 920 GB & \url{https://www.phonetik.uni-muenchen.de/forschung/Bas/SIGNUM/} &	P\\\cline{3-11}
\multirow{12}{*}{}& \multirow{3}{*}{} & RWTH-PHOENIX-Weather 2014 T &	Videos & - & 1,066 & 8,257 & Germany & 39 GB & \url{https://www-i6.informatik.rwth-aachen.de/~koller/RWTH-PHOENIX-2014-T/} &	P
\\\cline{2-11}
\multirow{12}{*}{}& \multirow{2}{*}{Greek} & GSL SD & Videos \& Depth from Real Sense	& 7 & 310 & 10,290 & Greece & -& \url{https://vcl.iti.gr/dataset/gsl/} &	CA\\\cline{3-11} \multirow{12}{*}{}& \multirow{2}{*}{} & GSL SI & Videos \& Depth from Real Sense &  7 & 310 &	10,290 & Greece & - & \url{https://vcl.iti.gr/dataset/gsl/} &	CA\\\cline{2-11} 
\multirow{12}{*}{}& \multirow{2}{*}{Korea } & KETI \cite{ko2019neural} & Videos	& 14 & 524 (419 words and 105 sentences) & 14,672 & South Korea & -& \url{https://arxiv.org/pdf/1811.11436.pdf} &	CA
\\\hline
\end{longtable}
\end{small}
\clearpage
\begin{center}
\begin{longtable}{ | p{0.8cm} | p{4.1cm} | p{7.2cm} | p{3.8cm} |}
\caption{SLR Datasets Vs. Modality. Highlight the related work in the literature regarding the various datasets and modalities. Dynamics and multi-modality like RGB, depth, and skeleton lead to better recognition rate in SLR.}
\label{Table 10}\\
\hline
\textbf{Year} & \textbf{Authors} & \textbf{Datasets} & \textbf{Modality} \\
\hline
2020 &	\citet{elboushaki2020multid}	& isoGD, 
SBU, NATOPS, SKIG &	RGB, Depth, Dynamic\\
\hline
2020 & \citet{rastgoo2020hand} &	RKS-PERSIANSIGN, NYU & \multirow{4}{*}{RGB, Dynamic} \\ 
\cline{1-3}
2019 &	\citet{kopuklu2019real} &	EgoGesture, NVIDIA benchmarks &\\ 
\cline{1-3}
2019 &	\citet{lim2019isolated} &	RWTH-BOSTON-50,
ASLLVD & \\ 
\cline{1-3}
2019 & \citet{chen2019construct}	& DHG-14/28 Dataset, SHREC’17 Track Dataset & \\ 
\hline
2019 & \citet{ferreira2019role} &	Real video samples	&  \multirow{3}{*} {RGB, Depth, Static} \\ \cline{1-3}
2019 & \citet{gomez2019accurate} &	STB &\\ 
\cline{1-3}
2018 & \citet{spurr2018cross} &	NYU, STB, MSRA, ICVL &\\ 
\cline{1-3}
2018 &	\citet{kazakos2018fusion} &	NYU & \\ 
\hline
2019 & \citet{li2019end} &	B2RGB-SH, STB &	\multirow{3}{*} {RGB, Static}\\ \cline{1-3}
2018 & \citet{mueller2018ganerated} &	EgoDexter, Dexter, STB &\\ 
\cline{1-3}
2017 &	\citet{victor2017handtrack} &	Egohands &\\ 
\hline
2018 & \citet{baek2018augmented}	& BigHand2.2M, MSRA, ICVL, NYU	&	\multirow{6}{*}{Depth, Static}\\ \cline{1-3}
2018	& \citet{moon2018v2v} &	MSRA, ICVL, NYU &\\ 
\cline{1-3}
2018 & \citet{ge2018robust} &	MSRA, ICVL, NYU &\\ 
\cline{1-3}
2017 & \citet{ge20173d}&	MSRA, NYU &\\ 
\cline{1-3}
2017 & \citet{dibra2017refine}&	ICVL, NYU &\\ 
\cline{1-3}
2016 & \citet{sinha2016deephand}&	NYU &\\ 
\hline
2017 & \citet{zimmermann2017learning} &	Dexter, STB &	3D, RGB\\
\hline
2018 &	\citet{marin20183d} &	UBC3V, ITOP	 & \multirow{4}{*}{3D, Depth}\\ \cline{1-3}
2017	 & \citet{deng2017hand3d} &	NYU &\\ 
\cline{1-3}
2016 & \citet{oberweger2016efficiently} &	MSRA &\\ 
\cline{1-3} 
2015 & \citet{oberweger2015hands} &	NYU &\\ 
\hline
2018 & \citet{rastgoo2018multi} &	Massey 2012, ASL Fingerspelling A, SL Surrey  & \multirow{2}{*}{2D, Depth, RGB}\\ \cline{1-3}
2016 &	\citet{duan2016multi} &	RGBD-HuDaAct,
isoGD&\\ 
\hline
2020 & \citet{chen1708pose} &	NYU, ICVL, MSRA & \multirow{10}{*}{2D, Depth}\\ \cline{1-3}
 2019 & \citet{dadashzadeh2019hgr}	& OUHANDS  &\\ 
\cline{1-3}
	2018 & \citet{wang2018drpose3d} &	Human3.6M  &\\ 
\cline{1-3} 
 2017	& \citet{yuan2017bighand2} &	BigHand2.2M, MSRA, ICVL, NYU  &\\ 
\cline{1-3}
2017 & \citet{guo2017towards} &	ITOP, MSRA, ICVL, NYU   &\\ 
\cline{1-3}
2017	& \citet{fang2017hand}&	ICVL, NYU  &\\ 
\cline{1-3}
2017	&  \citet{madadi2017end} &	MSRA, NYU  &\\ 
\cline{1-3}
2016 &	\citet{wang2016large} &	isoGD  &\\ 
\cline{1-3}
2016 & \citet{haque2016towards} &	EVAL, ITOP  &\\ 
\cline{1-3}
2015 &	\citet{tagliasacchi2015robust} &	Real video samples &\\ 
\hline
2020 &	\citet{rastgoo2020video}&	isoGD	& \multirow{5}{*}{2D, RGB}\\ \cline{1-3}
2016 &	\citet{wei2016convolutional}	& MPII, FLIC, LSP &\\ 
\cline{1-3}
2016 &	\citet{newell2016stacked}	& MPII, FLIC &\\ 
\cline{1-3}
2015 &	\citet{koller2015continuous} &	RWTH-PHOENIX-Weather &\\ 
\cline{1-3}
2014	& \citet{toshev2014deeppose} &	LSP, FLIC &\\ 
\hline
\end{longtable}
\end{center}
\begin{table}[]
\caption{Study of current state-of-the-art SLR model results with various datasets.}
\label{Table 11}
\begin{tabular}{ | p{2cm} | p{2cm} | p{0.5cm} | p{5.5cm} |p{5.5cm} |}
\hline
\textbf{State-of-the-art SLR  Model} &
  \textbf{Author} &
  \textbf{Year} &
  \textbf{Results} &
  \textbf{Datasets} \\ \hline
CoT4 CNN &
  \citet{ravi2019multi} &
  2019 &
  Recognition rate -89.69\% &
  BVCSL3D dataset \\ \hline
3 D CNN with score level fusion &
  \citet{gokcce2020score}&
  2020 &
  Accuracy   -94.94 \% &
  Bosphorus Sign22K \\ \hline
STMC &
  \citet{zhou2020spatial} &
  2020 &
  \begin{tabular}[c]{@{}l@{}}WER- 2.1 for Continuous SLR 100 dataset \\Split I case   and WER - 28.6 for  Split II case   \\ WER-20.7 for RWTH-PHOENIX-Weather 2014 \\WER - 21.0 for   RWTH-PHOENIX-Weather \\2014 T\end{tabular} &
  \begin{tabular}[c]{@{}l@{}}1. Continuous   SLR 100\\    \\ 2. RWTH-PHOENIX-Weather   2014\\    \\ 3. RWTH-PHOENIX- Weather 2014-T\end{tabular} \\ \hline
TK-3d convNet &
\citet{li2020transferring}  &
  2020 &
  \begin{tabular}[c]{@{}l@{}}Recognition accuracy - 77.55 \% for WLASL 100 \\and   68.75 \% for WLASL 200\\   Recognition accuracy - 83.91 \% for MSASL 100\\ and   81.14 \% for MSASL 200\end{tabular} &
  \begin{tabular}[c]{@{}l@{}}1. WLASL   100\\    \\ 2. WLASL   300\\    \\ 3. MSASL   100\\    \\ 4. MSASL   200\end{tabular} \\ \hline
SLRT &
  \citet{camgoz2020sign} &
  2020 &
  BLEU 4 scores -21.80 &
  RWTH-PHOENIX- Weather 2014-T \\ \hline
TSPNet –Joint &
 \citet{li2020tspnet} &
  2020 &
  BLEU 4 -13.41 &
  RWTH-PHOENIX- Weather 2014-T \\ \hline
Multi-stream   Conv Architecture &
 \citet{zheng2021enhancing}  &
  2021 &
  BLEU 4 score – 10.89 (RoI) and 10.73 (stream) &
  RWTH-PHOENIX- Weather 2014-T \\ \hline
DF- WiSLR (SVM augmented) &
  \citet{ahmed2020df} &
  2021 &
  Dynamic sign accuracy - 98.5 \% and Static sign accuracy   - 99.9 \% &
  Wi-Fi CSI dataset (49 gesture (static and dynamic) \\ \hline
SAN &
 \citet{slimane2021context} &
  2021 &
  WER – 29.78 \% &
  RWTH-PHOENIX-Weather 2014 \\ \hline
Inflated deep CNN &
  \citet{tongi2021application} &
  2021 &
  Accuracy -0.75 &
  SIGNUM \\ \hline
GLEN &
\citet{hu2021global} &
  2021 &
  Accuracy -69.9 \% for NMFs-CSL and accuracy - 96.8 \%   for Isolated SLR 500 dataset &
  \begin{tabular}[c]{@{}l@{}}1. NMFs-CSL\\    \\ 2. Isolated   SLR 500\end{tabular} \\ \hline
VTN-PF &
 \citet{de2021isolated} &
  2021 &
  Accuracy 92.92\% &
  AUTSL \\ \hline
SAM SLR &
  \citet{jiang2021skeleton} &
  2021 &
  Accuracy -98.42 \% for RGB and 98.53\% for RGB-D &
  AUTSL \\ \hline
SLRGAN &
  \citet{papastratis2021continuous} &
  2021 &
  \begin{tabular}[c]{@{}l@{}}Deaf-to-Deaf   SLRGAN WER - 36.05 for \\GSL SD and WER -2.26 for  GSL SI\\ \\SLRGAN  WER -  2.98 for GSL SI\\    \\ SLRGAN  WER - 37.11   for GSL SD\\    \\ WER - 23.4 \% for RWTH-PHOENIX- \\Weather 2014-T\\  \\ WER - 2.1 \% Continuous SLR 100\end{tabular} &
  \begin{tabular}[c]{@{}l@{}}1. GSL SD\\    \\ 2. GSL SI\\    \\ 3. RWTH-PHOENIX- Weather 2014-T\\    \\ 4. Continuous   SLR 100\end{tabular} \\ \hline
VMC &
  \citet{min2021visual} &
  2021 &
  \begin{tabular}[c]{@{}l@{}}WER - 1.6 \% for Continuous SLR 100\\    \\ WER -22.3 \% for RWTH-PHOENIX-\\Weather 2014\end{tabular} &
  \begin{tabular}[c]{@{}l@{}}1. Continuous   SLR 100\\    \\ 2. RWTH-PHOENIX-Weather   2014\end{tabular} \\ \hline
SMA-SLR- v2 &
 \citet{jiang2021sign}  &
  2021 &
  \begin{tabular}[c]{@{}l@{}}Accuracy -98.53 \% for AUSTL \\Accuracy - 59.39 \%   WLASL2000 dataset\\ per instance case and 56.63\% per class \\ Accuracy - 99 \% for Isolated SLR 500\end{tabular} &
  \begin{tabular}[c]{@{}l@{}}1. AUSTL \\    \\ 2. WLASL   200\\    \\ 3. Isolated   SLR 500\end{tabular} \\ \hline
SLR-Net- J+B &
  \citet{meng2021attention} &
  2021 &
  Accuracy -98.08 \% for Isolated SLR -500 and 64.57 \%   for DEVISIGN-L &
  \begin{tabular}[c]{@{}l@{}}1. Isolated   SLR -500\\    \\ 2. DEVISIGN-L\end{tabular} \\ \hline
SVM with RBFK (sEMG and acc) &
  \citet{pereira2022automatic} &
  2022 &
  Accuracy -96.66\% &
  Colombian sign language (3 subjects, 360 signs, 12   words) \\ \hline
SPOTER &
  \citet{bohavcek2022sign} &
  2022 &
  \begin{tabular}[c]{@{}l@{}}Accuracy – 100\% for LSA64 \\Accuracy -63.18\% for WLASL 100 and \\43.78 \% for WLASL   300\end{tabular} &
  \begin{tabular}[c]{@{}l@{}}1. LSA64\\    \\ 2. WLASL\end{tabular} \\ \hline
\end{tabular}
\end{table}
\twocolumn
\begin {figure}
\includegraphics[scale= 0.36]{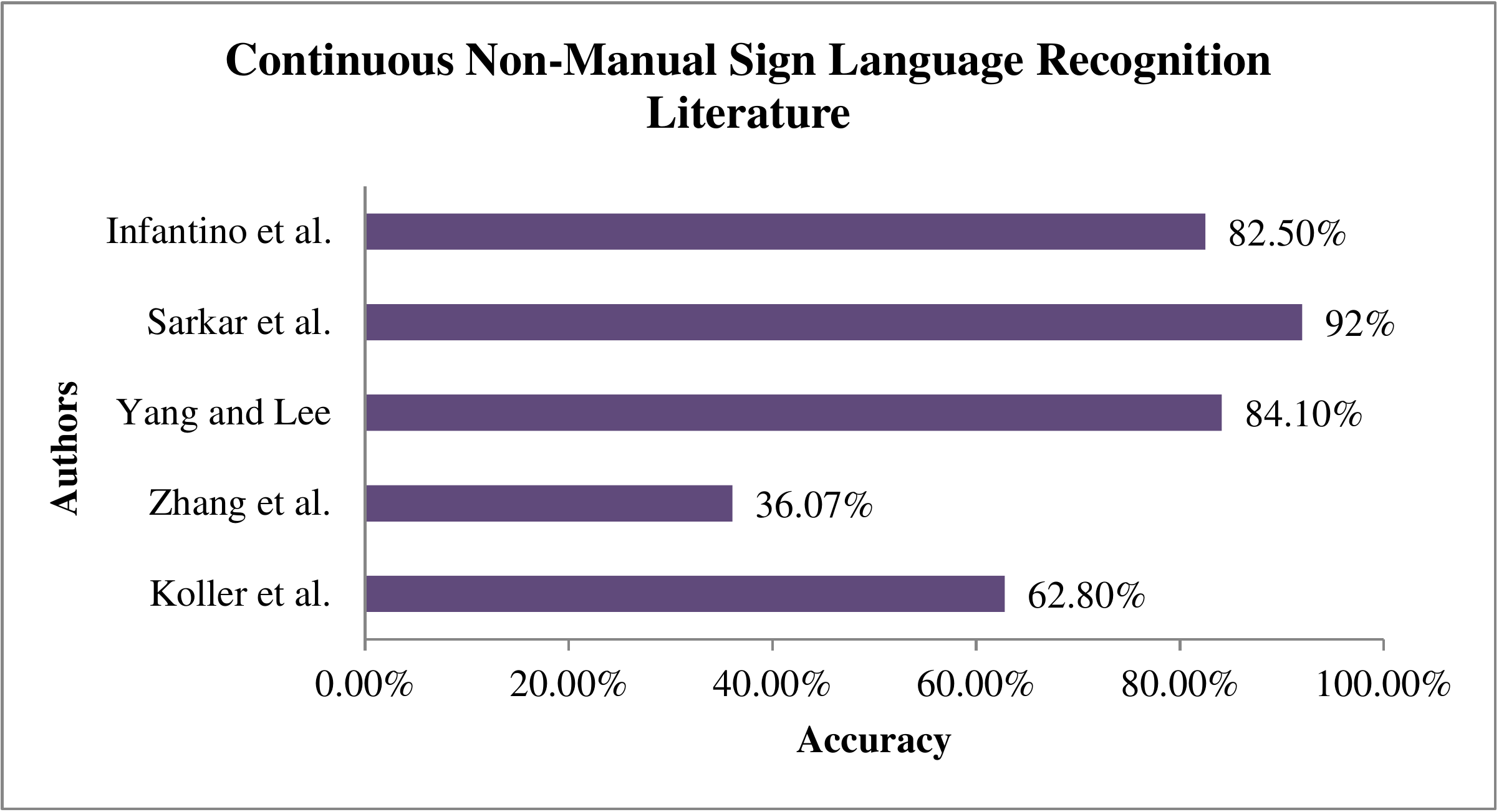}
\caption{Continuous Non-Manual SLR Literature. This bar chart representation shows that the literature's mean accuracy and recognition rate concerning the continuous non-manual SLR concerning different sign languages, models, and modalities.}
\centering
\label{Figure 10}
\end{figure}
\par \textbf{CNN and Hybrid methods:} \citet{koller2016deep} designed a continuous German SLR model using Iterative Expectation Maximization, incurred CNN. They trained the classifier with over a million hand shape sign data. \citet{brock2020recognition} performed Continuous Japanese SLR using CNN. They used frame-wise binary Random Forest for segmentation. The improvement of reliability, accuracy, and robustness for large-scale datasets could be a future research direction. \citet{zhou2020spatial} carried out a continuous SLR model using STMC (Spatial-Temporal Multi-Cue Network) to overcome the vision-based sequence learning problem. \citet{koller2020weakly} proposed a Hybrid CNN-LSTM-HMMs Continuous German SLR model. They performed sign language learning by sequential parallelism and validated it with three public sign language datasets.\\  
\par The continuous non-manual SLR-related research works are presented in tabular form in Table \ref{Table 7} and in graphical chart in Figure \ref{Figure 10} for better understating. The research on a continuous sign with a signer independent generic model is important because it has carried very little research on continuous SLR in the past decade.
\section{Classification Architectures} The classification is the brain of the SLR model. It aims to classify the sign accurately with minimum error. Researchers used various classifiers, e.g., traditional machine learning-based approach, deep learning-based approach, and hybrid approach.
\par ANN like back propagation, multi-layer, and recurrent neural networks are employed as classifiers, but handling large data is difficult. It requires enormous data for training to learn to challenge problems using a machine learning-based approach. The complication associated with HMM: 1. Likelihood of observation, 2. Best hidden state sequence decoding, 3. HMM, parameter framing. The parameter need for the 2 DCNN is excessively more, which makes the design process complex; this is the major drawback of 2 DCNN. In 3DCNN, the Spatio-temporal data has directly represented hierarchically, which is one of the unique features of 3 DCNN. Concerns to the long-term temporal dependence sign capturing 3 DCNN cannot assure robustness. LSTM eliminates the long-term dependence problem. The hybrid-based approach is adopted as a classifier to improve the accuracy. 
\begin {figure}
\includegraphics [scale=0.73]{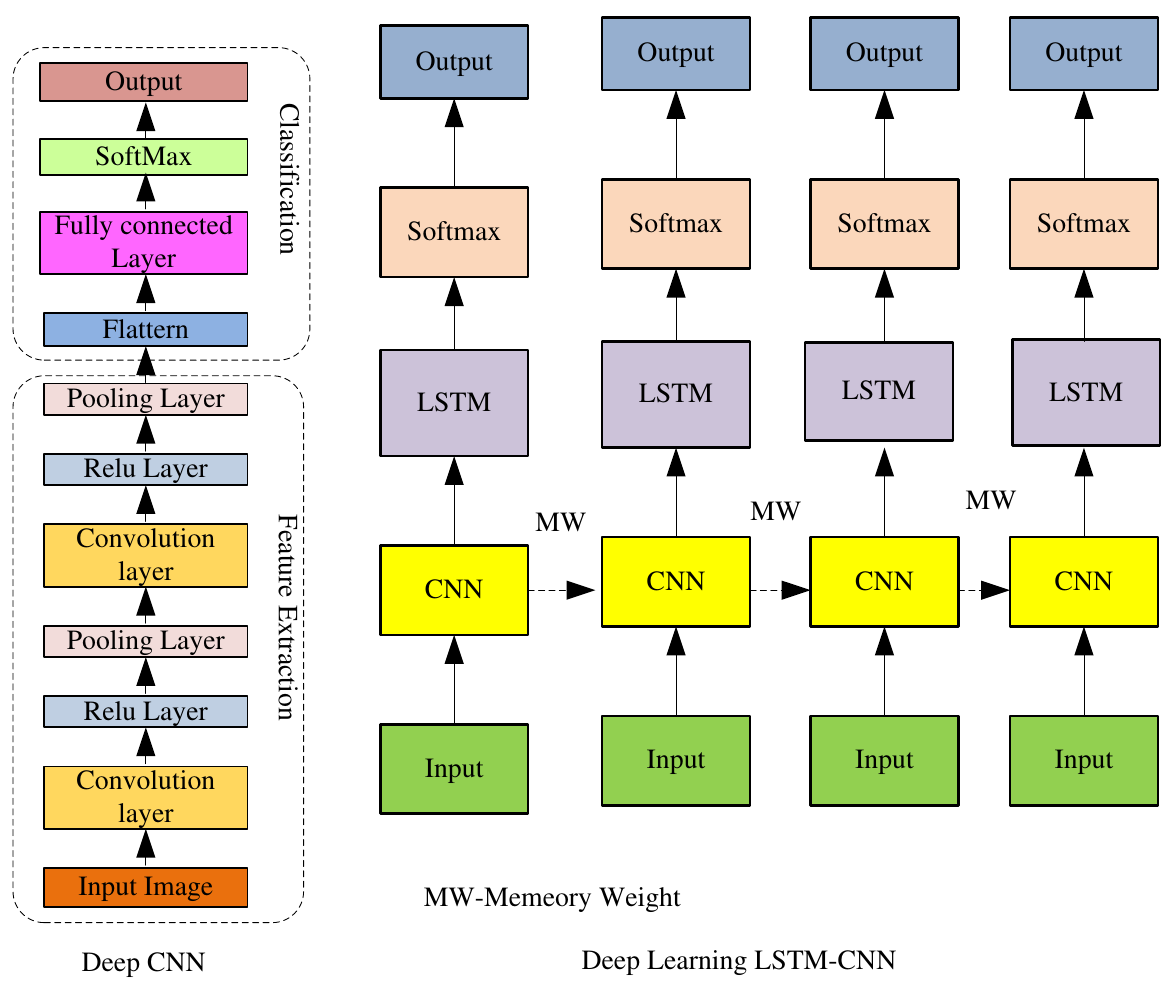}
\caption{Deep Learning Architectures (a) Deep CNN, (b) Deep Learning LSTM-CNN Architecture. The Deep CNN (3D) model and hybrid (LSTM+CNN) are recent classifier models. Hybrid deep learning-based classifier enhances recognition rate in SLR.}
\centering
\label{Figure 11}
\end{figure}

\subsection{Traditional Architectures} The Artificial Neural Network (ANN), Hidden Markov Model (HMM), and Recurrent Neural Network (RNN) are the most widely used classifiers in SLR due to their sequential data processing ability. \citet{fatmi2019comparing}, \citet{lee2021american}, \citet{von2008recent} carried the general ANN, RNN, and HMM-based SLR work. 
\subsection{Deep Learning Architectures} Deep learning makes massive growth in SLR recently. The spatial and temporal features are easy to handle by the deep learning models. LSTM can handle long-term dependence. Figure \ref{Figure 11} highlights the deep learning-based SLR architectures, namely Deep CNN and LSTM-CNN Architecture.
\subsection{Evaluation Metrics}
Computation of word error rate, accuracy, and recognition rate evaluates SLR models’ performance. The formulations used for evaluation are as follows:\\
\begin{equation}
Accuracy=(True Positive + True Negative)/(Total)      
\end{equation}
\begin{equation}
\begin{split}
WER (Word Error Rate) =(Number of substitutions+\\ Number of deletions + \\Number of insertions)/\\ (Total number of words in reference)
\end{split}
\end{equation}
\begin{equation}
\begin{split}
Recognition Rate = (Number of correctly identified \\images / Total Number of images)*100     
\end{split}
\end{equation}
The cross-validation scheme, namely leave-one-subject-out (LOSO) and k-fold cross-validation, is used to validate the SLR model's effectiveness. The Area Under the Curve (AUC) and ROC (Receiver Operating Characteristic) curve show the trade-off between true positive rate and false positive rate: it is used to measure the classifier performance. The Bilingual Evaluation Understudy (BLEU) score is used to measure the effectiveness of the translation.
\section{Different Types of Sensing Approach} According to the acquisition, it classifies SLR into two types, vision and sensor-based approaches. Many research works conducted both vision-based and sensor-based SLR to help the hand-talk community. Table \ref{Table 8} depicts sensing approach-based SLR existing work in literature.\\
\par \textbf{Vision-based sensing devices:} Types of cameras used for the vision-based approach are as follows:\\ Invasive device (body marker method): Examples: LED light, writ band, colored gloves.\\ Active devices: Kinect sensor, Leap motion sensor.\\ Stereo camera (depth camera): Capture depth information. \\ Single-camera: Smartphone, video camera, webcam, thermal camera, etc.\\ LMC (Leap Motion Controller): The LMC comprises three infrared LEDs and two cameras. It possesses the ability to track 850 nanometers' wavelength of light. The range is 60 cm (2 feet). Hand movement detection converts them into a suitable form of computer (commands) with the leap motion controller. Images are in a gray scalar format, and it acquired raw images using leap motion service software. Demerit: Accuracy is minimal.\\ Kinect Sensor: The skeleton (depth) image and movement creation done from three-dimensional image data. Multi-array microphone depth sensor and RGB camera are the comprised components in the Kinect sensor. Demerits: it requires more space (6 to 10 feet) distance between the sensor and signer.\\ 
\par \textbf{Sensor-based sensing devices:} The inexpensive and wearable sensor devices such as ACC (Accelerometer), Gyro (gyroscope), and sEMG (surface electromyogram) make sensor-based SLR a prominent tool for SLR.\\ 
Data Gloves (sensor-based): Analog form of signal converted into digital format by an ADC converter. It detects hand gestures and signs with the help of various sensors. It comprises an accelerometer and flex sensor (bend signal detection). A gyroscope gets orientation and angular and gains acceleration information with the help of an accelerometer. Finger bending information got by flex sensor. IMU (Inertial measurement units) used for hand movement estimation.\\ EMG (Electromyography): attach or insert the electrodes into the human muscle. With the help of the inserted electrodes, it recorded muscle movement as an electrical form. sEMG (surface electromyogram) is used for finger movement capturing and distinguishing.\\ The RF sensor possesses salient features that make it likable to acquire the sign in the dark environment, contact-less.\\ Radar and Wi-Fi: The motion of hand movements is collected from the multiple antennas. It extracts features based on Doppler shift and the difference between the magnitude. Advantage orientation and position are flexible.\\ It has captured the interaction between the environment background and the signer performing the sign using an ambient sensor. Example: 1. Temperature Sensor, 2. Radar Sensor, 3. Pressure Sensor, 4. Sound Sensor. 
\begin {figure}
\includegraphics [width=1.0\linewidth,height=0.3\textheight]{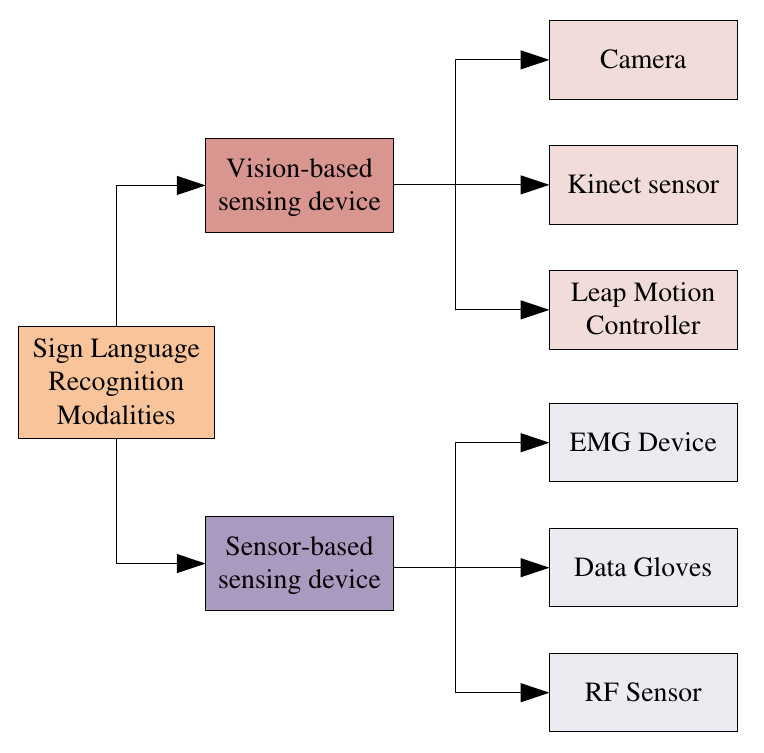}
\caption{SLR Modalities. It widely used Kinect and Leap Motion Controller sensor for vision-based SLR. EMG devices produce better results for sensor-based SLR, and it is an ongoing research area.}
\centering
\label{Figure 12}
  \end{figure}
In Figure \ref{Figure 12}, some SLR modalities are shown. 
\subsection{Various Datasets in SLR} There is a limitation on the datasets because most of the public datasets lack quality and quantity. The datasets are collected from native signers and ordinary people. Imitation of data was acquired to augment the dataset. We summarize the datasets available to SLR in Table \ref{Table 9}. The benchmark dataset details, including the URL link, are detailed here.
\subsubsection{Datasets Vs. Modality}
Datasets and modalities affect largely the performance of SLR models. Many researchers have implemented various SLR models using various methods and datasets. Table \ref{Table 10} shows Datasets Vs. Modality-based study concerns SLR. The RGB, depth, dynamic-based modality facilitates better performance. Hence, modality-based fusion leads to enhanced performance.
\subsubsection{The Complexities of SLR Datasets}
The acquisition of sign language is performed using the camera and sensor-based sensing devices like Kinect and leap motion controllers, armbands, gloves, electrodes (EMGs), etc. All datasets comprise its specific capturing format, modalities, mapping, environment, and illumination specific to the region or country sign language among 300 sign languages. Some of the complexities in these datasets are as follows:\\
\begin{itemize}
\item Acquire the SLR with various linguistic components based on multiple sensing devices is an arduous task, and the tedious process requires much time and effort.\\
\item Redundant and blur frames in data collection are a significant complexity that affects the SLR model recognition rate.\\
\item Complex background and various lighting are not considered in most datasets and have constraints; therefore, they cannot accurately recognize in a real-time application.\\
\item Most of the datasets were collected with a few native signers, only with a few repetitions. Therefore, it may not guarantee the signer independent recognition performance.\\
\item The signer wearing a long sleeve, occlusion, and object interaction during data collection makes it challenging to preprocessing and recognition.\\
\item The problem of handling real-time applications because most datasets are acquired using the constant background and illumination in a controlled environment.\\
\item Recognition of unseen sign words or sentences is difficult because the limited number of vocabularies and sentences present in the data make them incompatible in the real-time use case.\\
\end{itemize}
\subsubsection{The Solutions to Overcome the SLR Datasets Complexities}
The solutions to overcome the SLR datasets complexities are as follows:\\
\begin{itemize}
\item During sign data acquisition, the sample collection must consider various environments, lighting based on multiple times, performing the same words with different signers leads to sign independent SLR model and improves the generic ability.\\
\item The distance between the signers and the recording device should be feasible to overcome the blurring data issue.\\
\item Hand shape-based modality alone is not good enough to recognize the sign; thus, a non-manual feature-based dataset is required to perceive the grammar of the sign language. The isolated and continuous words/sentences include many signers and more repetitions based on a dataset with a more significant number of cues and corpus to improve accuracy, robustness, and generalization.\\
\item Versatile and massive corpus SLR dataset to address all sign components using multi-modal sensing with a complex and more extensive isolated, continuous sign without constraints based on capturing. Thus, it serves as a benchmark for SLR research to validate SLR model validity.\\
\end{itemize}
\subsection{Study of current state-of-the-art models for sign language recognition}
This paper further explores the state of the models presented in the sign language recognition as follows
\citet{ravi2019multi} performed Indian sign language recognition using RGB-D data using CNN models. They used four-stream inputs for training and tested performance on two streams (RGB spatial and temporal). They got a recognition accuracy rate of 89.69 \% for the BVCSL3D dataset.
\citet{gokcce2020score} carried out an isolated Turkish sign language recognition using 3 D residual CNN with score level fusion and got top 1 accuracy of 94.94 \% for the Bosphorus Sign22K dataset. 
\citet{li2020transferring} presented isolated sign language recognition using TK-3d convNet (transferring cross-domain knowledge-based 3D convolution network). Recognition accuracy of 77.55\% for WLASL 100 and 68.75\% for WLASL 200, 83.91\% for MSASL 100 and 81.14\% for MSASL 200 achieved based on the TK-3d convNet SLR model.
\citet{camgoz2020sign} SLRT (Sign language recognition and translation using transformer) applicability verified with RWTH-PHOENIX- Weather 2014-T dataset achieved 21.80 as BLEU 4 score.
\citet{li2020tspnet} suggested TSPNet – Temporal sematic pyramid network association of hierarchical feature learning based on continuous sign language recognition and result in BLEU 4 of 13.41 for RWTH-PHOENIX- Weather 2014-T dataset.
\par \citet{zheng2021enhancing} suggested a non-independent multi-stream convolutional and RoIs based multi-region convolutional architecture for sign language translation and obtained BLEU 4 scores – 10.89 (RoI) and 10.73 (stream) for RWTH-PHOENIX- Weather 2014-T.
\citet{ahmed2020df} presented Wi-Fi CSI (channel state information) dataset and developed sign language recognition using device-free Wi-Fi. SVM augmented-based model results with an accuracy of 98.5 \% for Dynamic sign and 99.9 \% for Static sign.
\citet{zhou2020spatial} designed a continuous sign language recognition based on STMC (spatial-temporal multi-cue network). They got the WER (word error rate) of 2.1, 28.6, 20.7, and 21.0 for Continuous SLR 100 dataset Split I case, Split II case, RWTH-PHOENIX-Weather 2014, and RWTH-PHOENIX-Weather 2014 T datasets, respectively.
\citet{slimane2021context} performed self-attention network (SAN-sign attention network)-based continuous sign language recognition. They used 2 D CNN with self-attention considered both hand and full-frame as inputs and combined to get final word glosses on evaluation on RWTH-PHOENIX-Weather 2014 dataset achieved WER of 29.78 \%.
\citet{tongi2021application} suggested the inflated deep CNN based on isolated SLR. They used the MSASL dataset to transfer the ASL knowledge to recognize GSL (German Sign Language) on the SIGNUM dataset and achieved an accuracy of 0.75 for high target data.
\citet{hu2021global} pointed out non-manual feature-aware GLEN (Global local enhancement network) based on the SLR model. They achieved a top 1 accuracy of 69.9\% for NMFs-CSL datasets and 96.8\% for isolated SLR 500 datasets.
\citet{de2021isolated} proposed Pose flow and hand cropping associated to video transformer network-based isolated sign language recognition. The VTN-PF (Video Transformer Network with hand cropping and pose) model evaluation on the AUTSL dataset got an accuracy of  92.92 \%.
\par \citet{jiang2021skeleton} devised a SAM SLR (Skeleton Aware multimodal framework Sign language recognition) concerning isolated sign language recognition. The skeleton-aware multi-modal (SSTCN–Separable spatial-temporal convolution network) results in better accuracy on the AUTSL dataset, with a top 1 accuracy of 98.42\% for RGB and 98.53\% for RGB RGB-D.
\citet{papastratis2021continuous} performed a generative adversarial network with transformer-based continuous sign language recognition. They used four datasets to validate the performance of SLRGAN (sign language recognition generative adversarial network). SLRGAN Deaf-to-Deaf SLRGAN achieves WER of 36.05 for GSL SD, WER of 2.26 for GSL SI, and  SLRGAN 
WER of 2.98 for GSL SI, WER of 37.11 for GSL SD, WER of 23.4\% for RWTH-PHOENIX- Weather 2014-T, and WER of 2.1\% Continuous SLR 100.
\citet{min2021visual} conducted VMC (visual alignment constraint) associated Resnet 18 backbone based on continuous sign language recognition model validated on Continuous SLR 100, and RWTH-PHOENIX-Weather 2014 datasets obtained a WER of  1.6\% and 22.3\%, respectively.
\par \citet{jiang2021sign} designed SMA-SLR- v2 (Skeleton aware multimodal framework with global ensemble model) based on isolated sign language recognition. They achieved the top 1 accuracy of 98.53\% for AUSTL (RGBD all), the top 1 accuracy of 59.39\% for the WLASL2000 dataset per instance case, and 56.63\% per class, and the top 1 accuracy of 99\% accuracy for isolated SLR 500 dataset.
\citet{meng2021attention} presented a GCN (graph convolution network)-based SLR network (dual sign language recognition model). The fusion of the two-stream models is SLR-Net-J+B, which results in the top 1 accuracy of 98.08\% for the isolated SLR -500 dataset and 64.57\% for the DEVISIGN-L dataset.
\citet{pereira2022automatic} devised Colombian sign language automatic recognition using SVM (support vector machine) with RBFK (radial basis function kernel) with four channels of sEMG (surface electromyography) and three-axis acc (accelerometer). Achieve accuracy 96.66\% for 12-word recognition.
\citet{bohavcek2022sign} performed isolated sign language recognition using SPOTER (Sign pose base transformer) validated with LSA64 and WLASL datasets, resulting in a  100\% accuracy for LSA64 and  63.18\% and 43.78\% accuracy for WLASL 100 and WLASL 300, respectively.
The current state-of-the-art SLR model is summarized in a Table \ref{Table 11} for better understanding. We hope this review paper sets a baseline for futuristic and advanced research in the SLR domain.  
\section{Discussion}
Sign language possesses dynamic gestures, trajectory property, and multi-dimensional feature vectors. These factors make it challenging to recognize sign language. Still, many researchers are attempting to develop a generalized, reliable, and robust SLR model. Multi-dimensional features are a novel approach that leads to a better recognition rate. This review paper aims to provide an easy understanding and helpful guidance to the research community. To perform research to develop an effective SLR model to assist the hand-talking community is one of the prominent domains in computer vision, pattern recognition, and natural language processing. 
\subsection{Limitation of Current Datasets and their sizes}
The ambiguities and lack of training dataset make the SLR vulnerable. Therefore, the standardized and large-scale datasets with manual and non-manual features are important. The limitation of the current datasets and their sizes are as follows:\\
Barrier concerning the recording/ collection/ measuring equipment:\\
\begin{itemize}
\item 	Poor camera quality affect the clarity of the sign in the vision-based system when the resolution is reduced leads to decreased accuracy.\\
\item Improper camera setup is another barrier because it leads to loss of important sign information when a sign is dynamic or static, performing the signer.\\
\item If a multi-camera set-up is used to acquire the signer data, the lack of synchronization lead to information loss, leading to poor performance.\\
\item	Device dependability should be reliable, cost-effective, and easy to maintenance.\\
\end{itemize}
The environment, background, and illumination profoundly affect the dataset preparation.\\
\begin{itemize}
\item When the background setup comprises noise, it creates misclassification and reduces the recognition rate, so it should be properly dealt with to overcome this barrier.\\
\item	Improper light and illumination reduce the clarity and also affect accuracy.\\
\item	The distance between the camera and the signer should be maintained at a nominal and workable range. Very much closer and farther, much long distance between the signer and the camera affect the performance.\\
\end{itemize}

\subsection{Limitation of Current Trends}
The limitations of the current trends in SLR are as follows:\\
\par The barrier regarding the different signers affects the accuracy:\\
\begin{itemize}
\item	Break off between the letter/ sign and speed up sign performing: The speedy, continuous, and frequent sign performed by the signer creates challenges for segmentation and feature extraction.\\
\item	Blockage of overlapping, occlusion of hand-face, hand-hand.\\
\item	Wearing a dress with long sleeves and wearing colored gloves also affects the sign recognition process.\\
\item	High variation concerns the interpersonal: Sign varies between signers and instants.\\
\end{itemize}
The barrier concerning the video domain:\\
\begin{itemize}
\item	The problem of handling the video data in the limited GPU memory is not tractable. Most CNN techniques are only image-based, videos that have an additional temporal dimension. A simple resizing process may cause a loss of crucial temporal information to perform the fine-tuning and classification process on each frame independently.\\
\end{itemize}
The barrier concerning network design in machine learning:\\
\begin{itemize}
\item	The recognition and classification ability prevailed by the location, illumination, and so on. \\
\item 	Higher batch size causes a fall in local convergence instead of global convergence. Smaller batch sizes lead to larger iterations and a rise in training expenses.\\
\item	Selection of the loss functions during training cause expenses.\\
\item	Selection of optimal hyperparameters. 
\end{itemize}
The active research domain is AI-based realistic modeling SLR translation and production of Avatar modeling (manual and non-manual). Developing AI Sign language learning and translation applications (web-based or smartphone) is one of the current trends. Although the advent of deep learning networks improves SLR accuracy, the limitations mentioned above still need to be addressed in the SLR domain.
\subsection{Other Potential Applications of SLR with Human-Computer Interaction}
Some potential applications of SLR with human-computer interaction are as follows:\\
\begin{itemize}
\item	Virtual reality: With the help of the electronics equipment, the user experiences artificial simulation of real world.\\
\item	Smart home: Home attributes to monitor, access, and control using artificial intelligence and electronics devices. It includes a security and alarm alert system. \\
\item	Health care: Intended to assist the patients in a better quality of life and good health care service. \\
\item	Social safety: To ensure safe and social engagement and to minimize social threats.  \\
\item	Telehealth: Remotely accessing clinical contacts and care services to enhance patients' health care. \\
\item	Virtual shopping: To provide hassle-free, more comfortable shopping with virtual stores.\\
\item	Digital signature: To transfer the information as an electronics sign.\\
\item 	Gaming and playing: To facilitate more entertaining, and gaming experience to users.\\
\item	Text and voice assistance: To provide better communication using technology and ease of user comfort.\\
\item	Education: To facilitate enhanced learning skills using advanced techniques.\\
\end{itemize}
\section{Future Direction and Research Scope}
Compared to the recent developmental achievement in automatic speech recognition, SLR is still lagging with a vast gap and remains at an early development stage. According to the literature study, a good number of research exists in SLR. Much research is struggling to achieve a high performance SLR model by exploring advanced techniques like deep learning, machine learning, optimization, and advanced hardware and sensor experimentation. Finally. We need a thorough exploration to solve the following issues in SLR.\\
\begin{itemize}
\item	Distinctiveness/contract of sign handling problem.\\
\item	Multiple sensors/camera fusion problems.\\
\item	Multi-modalities data handling issues.\\
\item	Computation problem.\\
\item	Consistency issues.\\
\item	Difficult to handle a large vocabulary.\\
\item	Requirement of standard datasets.\\
\end{itemize}
\textbf{Future Directions}\\
\par Future directions for SLR are as follows: 
\begin{itemize}
\item SLR model design needs a better understanding of optimal hyperparameter estimation strategy.\\ 
\item Building uncontrolled surrounding/environment-based SLR models is a thrust area because researchers develop most of the existing models in the literature with respect to the lab environment-based datasets. Hence, it is demanding. \\
\item Designing a user-friendly, realistic, and robust sign language model is one of the high-scope domains of SLR.\\
\item Design a high-precision sign language capturing device (sensor and camera).\\
\item Devise a novel training strategy to reduce computational training difficulty. \\
\item The lightweight CNN model for SLR is another research scope.\\
\item Develop an SLR with the association multi-modal based leverage to improve recognition accuracy. \\
\item Devising a generic automatic SLR model.\\
\end{itemize}
\par This review paper is presented to provide a complete guide to the research and allow the reader to know about the existing SLR works. It demonstrates challenging problems, research gaps, future research direction, and dataset resources. Therefore, the readers and researchers can move forward towards developing novel models and products to assist the hands-talk community and contribute to social benefits.
\section{Conclusion}
There are several review papers on hand gestures and SLR. Still, existing review papers do not comprehensively discuss facial expression, modality, and dataset-based sign language, lacking in-depth discussion. With this motivation, this review paper studied different types of SLR, various sensing approaches, modalities, and various SLR datasets and listed out the issues of SLR and the future direction of SLR. However, further complete guidance will provide a more precise understanding and acquire knowledge and awareness of the problem’s complexity, state-of-the-art models, and challenges in SLR. 
\par This comprehensive review paper will help to guide upcoming researchers about the SLR introduction, needs, applications, and processes involved in SLR. It discussed various manual and non-manual SLR models. Also, it reviewed isolated and continuous of each type (manual and non-manual) and provided easy understanding to the reader with the help table and diagram. The manual and non-manual type-based SLR present effectively and then examine the works related to the various modalities and datasets. Finally, we reviewed recent research progress, challenges, and barriers of existing SLR models, organized in an informative and valuable manner concerning the various types, modalities, and dataset. The improvisation of accuracy concerns vision-based SLR, one of the ongoing and hot research topics. The sensor-based approach is highly suitable for laboratory-based experimentation. But not an appropriate choice for practical real-time applications. The vision-based SLR model’s accuracy is less than the sensor-based approach and very much less than the speech recognition model. A robust and sophisticated method is essential for extracting manual and non-manual features and overcoming the barriers. Therefore, a lot of scopes are available to the SLR domain. We hope this review paves insight for readers and researchers to propose a state-of-the-art method that facilitates better communication and improves the hand-talk community’s human life.
\\
\section*{Declaration}
\noindent\textbf{Funding-}
{No Funding was received for this work.} \\
\textbf{Competing interests-}
{The authors declare no conflict of interest.}\\
\textbf{Availability of data and materials-}
{Not Applicable.}\\
\textbf{Code availability-}
{Not Applicable.}\\
\section*{Acknowledgment}
This work of Dr. M. Madhiarasan was supported by the MHRD (Grant No. OH-31-24-200-328) project.
\ifCLASSOPTIONcaptionsoff
\newpage
\fi
\bibliography{IEEEabrv,bibtex/bib/paper}
%
\newpage
\section*{Authors Biography}

%
\vspace{-9cm}
\begin{IEEEbiography}[{\includegraphics[width=1in,height=1.25in,clip,keepaspectratio]{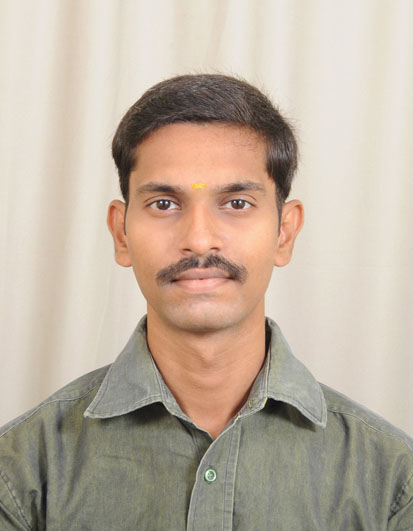}}] {Dr. M. MADHIARASAN} (Member, IEEE) has completed his Bachelor of Engineering degree in Electrical and Electronics Engineering in the year 2010 from Jaya Engineering College, Thiruninravur, under Anna University, Tamil Nadu, India, Master of Engineering degree in Electrical Drives and Embedded Control (Electrical Engineering) in the year 2013 from Anna University, Regional Centre, Coimbatore, under Anna University, Tamil Nadu, India and Ph.D. (Electrical Engineering) in the year 2018 from Anna University, Tamil Nadu, India. \\
He has worked as an Assistant Professor and R \& D In-charge in the Department of Electrical and Electronics Engineering, Bharat Institute of Engineering and Technology, Hyderabad, India, from 2018 to 2020. He is presently working as a Post-Doctoral Fellow at the Department of Computer Science and Engineering, Indian Institute of Technology, Roorkee (IITR). \\
His research areas include Renewable Energy Systems, Power Electronics and Control, Computer Vision, Human-Computer Interface, Pattern Recognition, Artificial Intelligence, Neural Networks, Optimization, Machine Learning, Deep Learning, Soft Computing, Internet of Things and Modeling and Simulation. He has been a Technical Program Committee Member, International Scientific Committee Member of many international conferences. He acts as an editorial board member and reviewer for many peer-reviewed international journals (Springer, Elsevier, IEEE Access, etc.) and a guest editor of Energy Engineering. \\
\end{IEEEbiography}
\vspace{-9cm}
\begin{IEEEbiography}[{\includegraphics[width=1in,height=1.25in,clip,keepaspectratio]{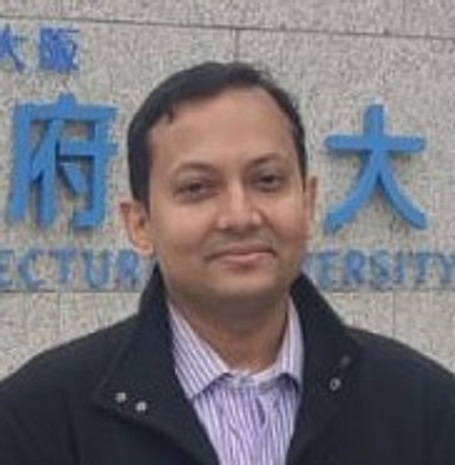}}]{Dr. PARTHA PRATIM ROY} (Member, IEEE) was a Postdoctoral Research Fellow with the RFAI Laboratory, France, in 2012, and the Synchromedia Laboratory, Canada, in 2013. He was with the Advanced Technology Group, Samsung Research Institute, Noida, India, from 2013 to 2014. He is currently an Associate Professor with the Department of Computer Science and Engineering, IIT Roorkee, India. He has authored or coauthored more than 200 articles in international journals and conferences. His research interests are pattern recognition, bio-signal analysis, EEG-based pattern analysis, and multilingual text recognition. He is an Associate Editor of the IET Image Processing, IET Biometrics, IEICE Transactions on Information and Systems and Springer Nature Computer Science.\\
\end{IEEEbiography}
\end{document}